\documentclass{article}

\usepackage[preprint]{neurips_2024}

\usepackage[utf8]{inputenc} 
\usepackage[T1]{fontenc}    
\usepackage{hyperref}       
\usepackage{url}            
\usepackage{booktabs}       
\usepackage{amsfonts}       
\usepackage{nicefrac}       
\usepackage{microtype}      
\usepackage{xcolor}         
\usepackage{lipsum}
\usepackage{fancyhdr}       
\usepackage{graphicx}       
\graphicspath{{media/}}     
\usepackage{xcolor}
\usepackage{longtable}
\usepackage{float}
\usepackage{authblk}
\usepackage{subcaption}
\usepackage{tcolorbox}
\usepackage{algorithm}
\usepackage{algorithmic}
\usepackage{xcolor}
\usepackage{listings}
\usepackage{ulem}
\usepackage{enumitem}
\usepackage{wrapfig}
\usepackage{tabularx}
\usepackage{bbm}
\usepackage{multirow}
\usepackage{colortbl}
\usepackage{caption}
\usepackage{natbib}
\setcitestyle{numbers,square}
\usepackage{amsmath}
\usepackage{tikz}
\newcommand*\circled[1]{\tikz[baseline=(char.base)]{
            \node[shape=circle,draw,inner sep=0pt] (char) {#1};}}

\definecolor{mygreen}{HTML}{3cb44b}
\definecolor{myred}{HTML}{ff3333}
\definecolor{Gray}{gray}{0.93}
\definecolor{shapecolor}{rgb}{0.0,0.5,0.0}

\usepackage{xcolor}

\newcommand{\dagnote}[1]{$\text{#1}^\dag$}

\newcommand{\model}{\textsc{MixLoRA}}

\newcommand{\doramodel}{\textsc{MixDoRA}}

\title{\model{}: Enhancing Large Language Models Fine-Tuning with LoRA-based Mixture of Experts}

\author[*]{Dengchun Li}
\author[*]{Yingzi Ma}
\author[*]{Naizheng Wang}
\author[*]{Zhengmao Ye}
\author[1]{Zhiyuan Cheng}
\author[*]{Yinghao Tang}
\author[3]{Yan Zhang}
\author[*]{Lei Duan}
\author[*]{Jie Zuo}
\author[2]{Cal Yang}
\author[*]{Mingjie Tang}
\affil[*]{\it Sichuan University, Chengdu, China}
\affil[1]{\it Purdue University, West Lafayette, USA}
\affil[2]{\it Emory University, Atlanta, USA}
\affil[3]{\it Nanyang Technological University, Singapore}
\affil[ ]{\tt
mikecovlee@163.com\\
\{g19myz, pherenice1125, yezhengmaolove\}@gmail.com\\
cheng443@purdue.edu,\\
\{yinghaotang2001, yanzhang.jlu\}@gmail.com,\\
\{leiduan, zuojie\}@scu.edu.cn,\\
j.carlyang@emory.edu, tangrock@gmail.com}

\begin{document}

\maketitle

\begin{abstract}

Fine-tuning Large Language Models (LLMs) is a common practice to adapt pre-trained models for specific applications. While methods like LoRA have effectively addressed GPU memory constraints during fine-tuning, their performance often falls short, especially in multi-task scenarios. In contrast, Mixture-of-Expert (MoE) models, such as Mixtral 8x7B, demonstrate remarkable performance in multi-task learning scenarios while maintaining a reduced parameter count. However, the resource requirements of these MoEs remain challenging, particularly for consumer-grade GPUs with less than 24GB memory. To tackle these challenges, we propose \model{}, an approach to construct a resource-efficient sparse MoE model based on LoRA. \model{} inserts multiple LoRA-based experts within the feed-forward network block of a frozen pre-trained dense model and employs a commonly used top-k router. Unlike other LoRA-based MoE methods, \model{} enhances model performance by utilizing independent attention-layer LoRA adapters. Additionally, an auxiliary load balance loss is employed to address the imbalance problem of the router. Our evaluations show that  \model{} improves about 9\% accuracy compared to state-of-the-art PEFT methods in multi-task learning scenarios. We also propose a new high-throughput framework to alleviate the computation and memory bottlenecks during the training and inference of MOE models. This framework reduces GPU memory consumption by 40\% and token computation latency by 30\% during both training and inference.

\end{abstract}

\section{Introduction} \label{sec:intro}


Instruction fine-tuning of Large Language Models~(LLMs)~\cite{Brown2020LanguageMA, Chowdhery2022PaLMSL, Hoffmann2022TrainingCL, Touvron2023LLaMAOA, Touvron2023Llama2O} for various downstream tasks has achieved impressive proficiency in Natural Language Processing~(NLP)~\cite{Chung2022ScalingIL, Iyer2022OPTIMLSL, zheng2023judging}. As the scale of parameters increases, LLMs have been demonstrated to be able to identify complex linguistic patterns, thereby enabling the emergence of powerful cross-task generalization capabilities~\cite{wei2022emergent}. The paradigm of instruction tuning leads to a trade-off between the computational resources required and the performance achieved on downstream tasks, which has been a valuable facet.


\begin{figure}[!t]
    \centering
    \includegraphics[width=1\linewidth, trim= 0 0 0 0, clip]{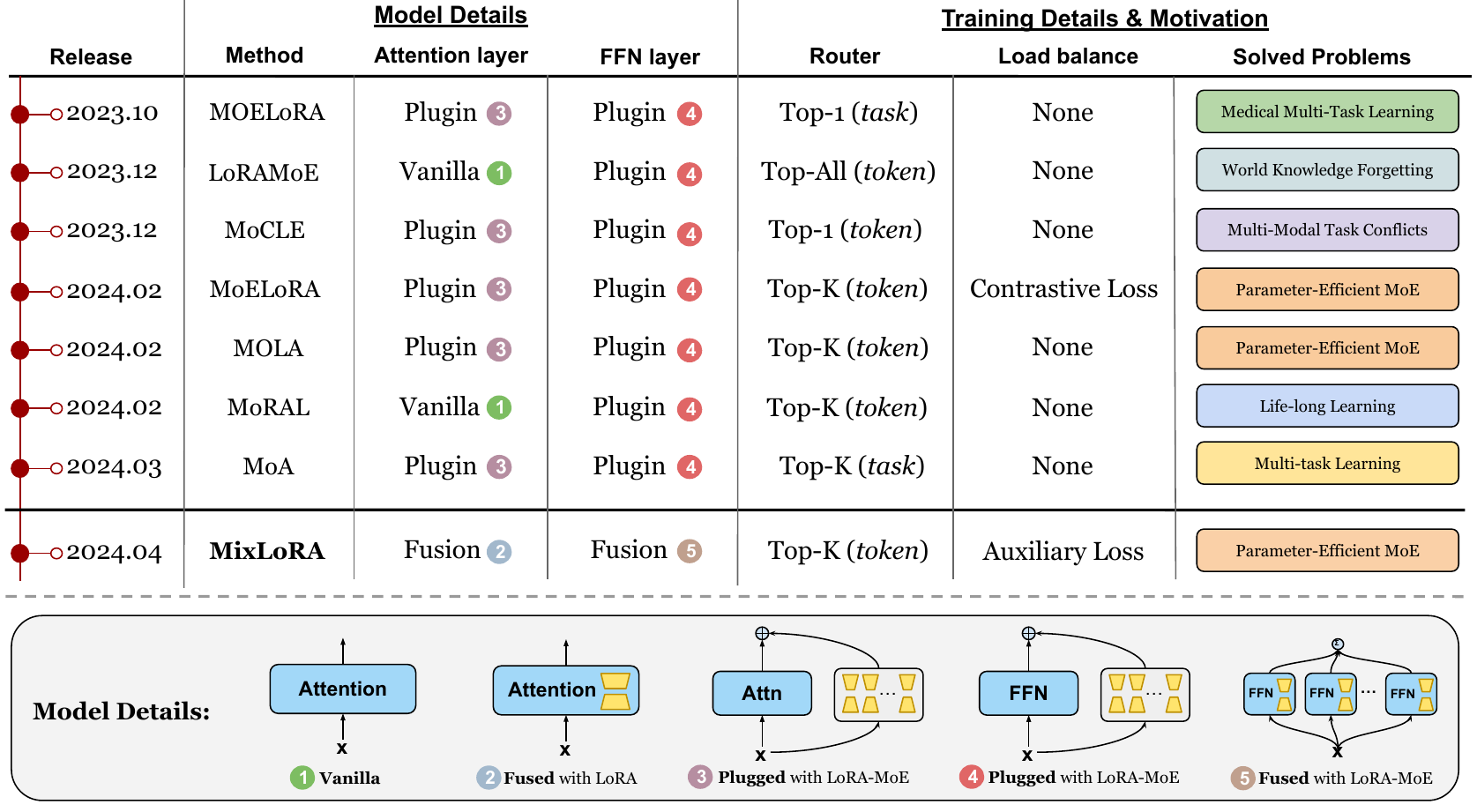}
    \caption{The timeline of public LoRA-MoE methods' release dates, including the detailed model information on the position of integration, how to train with the LoRA-MoE method (router and load balance), and the problems they aim to solve.}
    \label{fig:lora_moe_compare}
\end{figure}

To substantially reduce the computational and memory resources required by full parameter fine-tuning processes, Parameter-Efficient Fine-Tuning (PEFT) methods have emerged~\cite{houlsby2019parameter,Li2021PrefixTuningOC,Lester2021ThePO,BenZaken2021BitFitSP,Liu2022FewShotPF,Hu2021LoRALA}. Among these methods, Low-Rank Adaptation (LoRA)~\cite{Hu2021LoRALA}, a popular PEFT approach, offers performance comparable to full fine-tuning across various downstream tasks while demanding less computational effort. This is achieved by introducing low-rank adaption matrices to simulate the gradient updates while keeping pre-trained model weights frozen. However, its performance still falls short of full fine-tuning~\cite{sun2022recent, sundaram2019survey}. Recent studies, such as LoRA+~\cite{hayou2024lora+} and DoRA~\cite{liu2024dora}, aim to enhance the performance of LoRA by optimizing the parameter updating process. Despite these improvements, LoRA-based methods face challenges in handling multiple tasks simultaneously due to limited trainable parameters and the problem of catastrophic forgetting~\cite{feng2024mixtureofloras, huang2024lorahub}, which hampers cross-task generalization in LLMs. To mitigate this, some approaches attempt to integrate additional components into existing methods. For instance, AdapterFusion~\cite{pfeiffer2020adapterfusion} and LoRAHub~\cite{huang2024lorahub} introduce task-specific LoRA modules or experts, combining knowledge into domain adapters through additional attention fusion and element-wise composition. However, these additional components often add complexity to the training process, limiting their applicability compared to simpler PEFT methods like LoRA and its variants.

A promising solution is to design an architecture that combines LoRA's resource-saving features with the versatility of MoE models. This approach involves adding multiple LoRA modules as experts in transformer sublayers (i.e., attention and feedforward layers)~\cite{yang2024moral,luo2024moelora,feng2024mixtureofloras}, using a router to assign experts to tokens~\cite{yang2024moral,wu2024parameterefficient,dou2024loramoe,gou2024mixture,Liu2023MOELoRAAM,feng2024mixtureofloras}, and incorporating an expert balancing loss to prevent uneven token distribution~\cite{yang2024moral,wu2024parameterefficient}. As shown in Fig. \ref{fig:lora_moe_compare}, most methods in this direction focus on solving specific-domain problems such as medical multi-task learning~\cite{Liu2023MOELoRAAM} and world knowledge forgetting~\cite{dou2024loramoe}. Only MoELoRA~\cite{luo2024moelora} and MOLA~\cite{gao2024higher} are proposed to enhance the general capability of multi-task learning for LLMs. However, they only focus on single-task learning while ignoring multi-task learning, leading to certain limitations. We observe that these LoRA-MoE models plug multiple LoRA modules into a single self-attention or FFN block to form the MoE structure (\circled{3} and \circled{4} in Figure~\ref{fig:lora_moe_compare}), while the high-performance pre-trained MoE models, such as Mixtral 8x7B~\cite{jiang2024mixtral}, often use multiple FFN based expert networks during forward propagation. Moreover, research on vanilla transformers~\cite{vaswani2023attention} indicates that employing MoE only on the FFN block can be more efficient than on both self-attention and FFN blocks, with fine-tuning the attention layer with LoRA can further enhance MoE models~\cite{zoph2022stmoe}. Additionally, these LoRA-MoE methods introduce multiple LoRAs within a single transformer block, providing opportunities to improve the computational efficiency using Multi-LoRA parallel computing techniques~\cite{ye2023aspen}.


Inspired by these, we propose \model{}, which fuses multiple LoRAs with the shared FFN layer and employs them to store updated parameters for each expert during fine-tuning, thereby aligning it more with conventional MoE models such as Mixtral 8x7B~\cite{jiang2024mixtral}. Additionally, we employ LoRA, instead of MoE, on the self-attention layer to further improve the performance of \model{}. Following the previous work~\cite{fedus2022switch}, we also employ an auxiliary load balance loss for \model{} to address the expert imbalance problem. Furthermore, we design a high-performance framework to optimize the computation process of multiple LoRA based experts in \model{} both for training and inference\footnote{GitHub: \url{https://github.com/TUDB-Labs/MixLoRA}}. As a result, this framework reduces the computational complexity of \model{} by 30\%, and saves about 40\% GPU memory usage when training or inferencing multiple \model{} models. \looseness=-1

We validate the effectiveness of \model{} across a wide variety of tasks. For baselines, we choose the widely used PEFT method LoRA~\cite{Hu2021LoRALA} and the current state-of-the-art PEFT method DoRA~\cite{liu2024dora}. Quantitative results demonstrate that \model{} consistently outperforms LoRA and DoRA in both single-task and multi-task learning scenarios. For single-task learning, \model{} achieves an average accuracy improvement of \textbf{5.8\%} compared to LoRA and \textbf{2.3\%} compared to DoRA on LLaMA-2 7B. In multi-task learning, \model{} significantly surpasses LoRA by \textbf{9.8\%} and DoRA by \textbf{9\%} in accuracy, while demonstrating less performance degradation compared to the baseline methods. In summary, our contributions in this paper are as follows:
\begin{enumerate}[wide, labelwidth=0pt, labelindent=0pt]
    \item We introduce \model{}, a parameter-efficient mixture-of-experts method that constructs multiple LoRA based experts and a frozen shared FFN block from a pre-trained dense model. This approach simplifies the creation of an efficient sparse mixture-of-experts model with limited resources, and enhances performance across various downstream tasks.
    \item We implement a high-throughput framework for the training and inference process. By optimizing redundant overhead in the computation process, we achieve approximately a \textbf{30\%} reduction in the token computation latency of \model{}, and save \textbf{40\%} or more in GPU memory usage during both training and inference of multiple \model{} models on a single consumer-grade GPU with 24GB memory, using LLaMA2-7B in half precision.
    \item We conduct comprehensive evaluations on several commonly used benchmarks, including ARC~\cite{clark2018think}, BoolQ~\cite{clark2019boolq}, OpenBookQA~\cite{mihaylov2018suit}, PIQA~\cite{bisk2020piqa}, SIQA~\cite{sap2019socialiqa}, HellaSwag~\cite{zellers2019hellaswag} and WinoGrande~\cite{sakaguchi2021winogrande}. The results demonstrate that \model{} exhibits superior performance in handling various downstream tasks compared to existing fine-tuning methods. Specifically, \model{} increase the average accuracy by \textbf{5.8\%} on single-task evaluations and \textbf{9.8\%} on multi-task evaluations for LoRA with LLaMA2-7B.
\end{enumerate}

\section{Related Works} \label{sec:related}
\noindent \textbf{PEFT For LLMs.}  Large Language Models~(LLMs)\cite{Brown2020LanguageMA, Chowdhery2022PaLMSL, Hoffmann2022TrainingCL, Touvron2023LLaMAOA, Touvron2023Llama2O} have demonstrated remarkable capabilities in NLP tasks.  Following this advancement, instruction finetuning~\cite{Chung2022ScalingIL, Iyer2022OPTIMLSL, zheng2023judging} has further enabled LLMs to understand human intentions and follow instructions, serving as the foundation of chat systems~\cite{chatgpt, gpt4}. However, as the size of LLMs scales up, finetuning them becomes a process that is time-consuming, and memory-intensive. To mitigate this issue, various studies explore different approaches: parameter-efficient finetuning~(PEFT)\cite{peft}, distillation\cite{Liu2023LLMQATDQ, Xiao2023OffsiteTuningTL}, quantization~\cite{Frantar2022GPTQAP, Xiao2022SmoothQuantAA}, pruning~\cite{Frantar2023SparseGPTML, Ma2023LLMPrunerOT}, etc. LoRA~\cite{Hu2021LoRALA}, leveraging low-rank matrices to decompose linear layer weights is one of the most popular PEFT methods, thus enhancing model performance without introducing any additional computational overhead during inference. For instance,  VeRA~\cite{kopiczko2024vera} incorporate learnable scaling vectors to adjust shared pairs of frozen random matrices across layers. Furthermore, FedPara~\cite{hyeonwoo2023fedpara} concentrates on low-rank Hadamard products for federated learning scenarios. Tied-Lora~\cite{renduchintala2023tiedlora} implement weight tying to further reduce the number of trainable parameters. AdaLoRA~\cite{zhang2023adalora} employ Singular Value Decomposition (SVD) to decompose matrices and prune less significant singular values for streamlined updates. DoRA~\cite{liu2024dora} decomposes pre-trained weights into two components, magnitude and direction, and utilizes LoRA for directional updates during fine-tuning, thereby efficiently reducing the number of trainable parameters.

\noindent \textbf{Mixture-of-Experts.} The concept of Mixture-of-Experts (MoE)~\cite{Jacobs1991AdaptiveMO} dates back to as early as 1991, introducing a novel supervised learning approach involving multiple networks (experts), each specialized in handling a subset of training examples. 
The modern incarnation of MoE modifies the traditional feed-forward sub-layer within transformer blocks by incorporating sparsely activated experts, thereby enabling substantial increases in model width without a corresponding surge in computational demands. Various MoE architectures have emerged, distinguished by their sampling strategies and routing mechanisms. Building on this evolution, LLaVA-MoLE~\cite{chen2024llavamole} effectively routes tokens to domain-specific experts within transformer layers, mitigating data conflicts and achieving consistent performance gains over plain LoRA baselines. For other MoE based methods, MoRAL~\cite{yang2024moral}addresses the challenge of adapting LLMs to new domains/tasks and enabling them to be efficient lifelong learners. LoRAMoE~\cite{dou2024loramoe} integrates LoRAs using a router network to alleviate world knowledge forgetting. PESC~\cite{wu2024parameterefficient} transitions dense models to sparse models using a MoE architecture, reducing computational costs and GPU memory requirements. MoE-LoRA~\cite{luo2024moelora} propose a novel parameter-efficient MoE method with Layer-wise Expert Allocation (MoLA) for Transformer-based models. MoCLE~\cite{gou2024mixture} proposes a MoE architecture to activate task-customized model parameters based on instruction clusters. As for ours, we integrate LoRAs as stochastic experts, reducing computational cost while expanding model capacity and enhancing LLMs' generalization ability. \looseness=-1

\begin{figure}[!t]
    \centering
    \includegraphics[width=0.9\linewidth, trim= 0 0 0 0, clip]{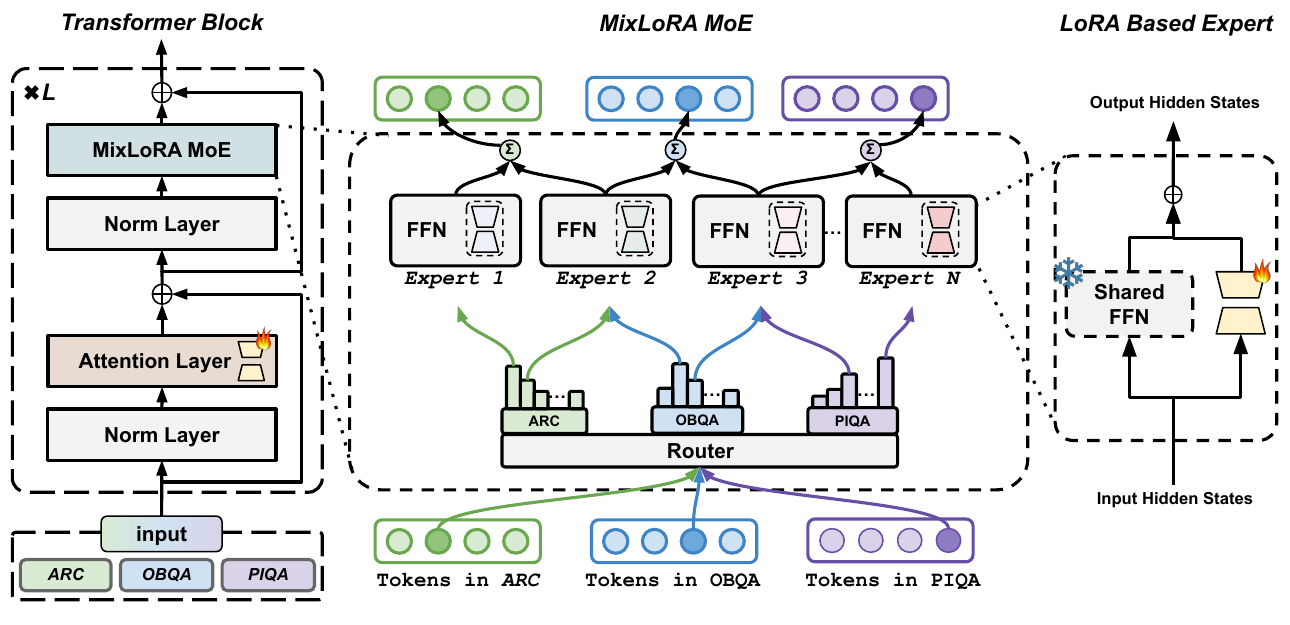}
    \caption{The architecture of \model{} transformer block. \model{} consists of n experts formed by an original FFN sublayer combined with different LoRAs, where the weights of the FFN sublayer are shared among all experts.}
    \label{fig:mixlora}
\end{figure}

\section{\model} \label{sec:method}


In this  section, we first introduce the foundational concepts of LoRA and MoEs in Section~\ref{sec:preliminary} and the architectural design of \model{} in Section~\ref{sec:arch}. Then,  we explain how we optimize the performance of \model{} for model training and inference in Section~\ref{sec:optimization},. 

\subsection{Preliminaries} \label{sec:preliminary}
\paragraph{LoRA.} LoRA only tunes the additional adaptation parameters and replaces the original weight update. A LoRA block consists of two matrices, $\mathbf{B} \in \mathbb{R}^{d_1 \times r}$ and $\mathbf{A} \in \mathbb{R}^{r \times d_2}$, where $d_1$ and $d_2$ denote the dimensions of the pretrained weights $\mathbf{W}$ of the LLMs ($\mathbf{W} \in \mathbb{R}^{d_1\times d_2}$). The parameter $r$ denotes the hidden dimension of Lora, with $r \ll min(d_1, d_2)$. Then, the updated weights $\mathbf{W^{'}}$ are calculated through:
\begin{equation}
    \boldsymbol{W}' = \boldsymbol{W} + \boldsymbol{B}\boldsymbol{A}
\label{eq:adapter}
\end{equation}

\paragraph{Mixture-of-Experts~(MoE).} The MoE architecture is initially introduced to language models through GShard~\cite{lepikhingshard}. An MoE layer consists of $n$ experts, denoted by $\{E_i\}_{i=1}^{n}$, along with a router $R$. The output $\mathbf{h^{'}}$ of an MoE layer for a given hidden states $\mathbf{h}$ is determined by: 
\begin{equation}
\mathbf{h^{'}} = \sum_{i=1}^{n} R(\mathbf{h})_iE_i(\mathbf{h}).
\label{eq:moe}
\end{equation}
Here, $R(\mathbf{h})_i$ indicates the router's output for the $i$-th expert for selecting specific experts, and $E_i(\mathbf{h})$ is the result from the $i$-th expert.

\subsection{Architecture of \model} \label{sec:arch}



As shown in Figure~\ref{fig:mixlora}, \model{} is constructed from two main parts. The first part involves constructing the sparse MoE block using the vanilla transformer block augmented with LoRAs. The second part utilizes a top-k router to assign each token from various tasks~(such as ARC, OBQA, PIQA, etc.) to different expert modules. Given the input text is $\boldsymbol{s}=\left(s_1,s_2,\dots,s_n\right)$ with label of $\boldsymbol{y}$. Let $\mathbf{h}_{i}^{\ell}\in\mathbb{R}^{1\times d}$~($1\leq i \leq n, 1 \leq \ell \leq L$) denote the output hidden state of the $i$-th token at the $\ell$-th large language model~(LLM) layer, where $L$ is the total number of the LLM layers and $d$ is the hidden dimension. The large language model consists of stacked multi-head self-attention (MSA) and feed-forward neural networks (FFN). Layer normalization (LN) and residual connections are applied within each block~\cite{wang2019learning, baevski2018adaptive}. Formally, the output $\mathbf{h}^{\ell}$ of the $\ell$-th LLM layer in a normal transformers block is calculated via:

\vspace{-0.3cm}
\begin{equation}
    \mathbf{h}^{0} =[s_{1},s_{2},\cdots,s_{n}], 
\end{equation}
\vspace{-0.3cm}
\begin{equation}
    \mathbf{z}^{\ell} =\text{MSA}(\text{LN}(\mathbf{h}^{\ell-1}))+\mathbf{h}^{\ell-1}, \quad \mathbf{h}^{\ell} =\text{FFN}(\text{LN}(\mathbf{z}^{\ell}))+\mathbf{z}^{\ell}
\label{ffn2}
\end{equation}



\paragraph{\model{} Forward.} \model{} constructs experts based on the LoRA technique \cite{Hu2021LoRALA}. \model{} utilizes LoRA to effectively store the updated parameters of each expert during fine-tuning, rather than employing LoRA solely for constructing each expert. This approach aligns \model{} more closely with existing pre-trained MoE models. In \model{}'s MoE blocks, the base weights of these experts are shared from a single feed-forward network (FFN) of the dense model (as illustrated in Figure~\ref{fig:mixlora}) to enhance both training and inference efficiency:

\begin{equation}
    \mathbf{h}^{\ell} =\text{MixLoRA}(\text{LN}(\mathbf{z}^{\ell}))+\mathbf{z}^{\ell}
\label{mixloraffn}
\end{equation}
\begin{equation}
\text{MixLoRA}(\mathbf{h}^{\ell}) = \sum_{k=1}^{K} R^{\ell}(\mathbf{h}^{\ell})_kE^{\ell}_{k}(\mathbf{h}^{\ell}), \quad E^{\ell}_{k}(\mathbf{h}^{\ell}) = \boldsymbol{W}^{\ell}\cdot\mathbf{h}^{\ell} + \boldsymbol{B}^{\ell}_{i}\boldsymbol{A}^{\ell}_{i}\cdot\mathbf{h}^{\ell}
\label{mixlora1}
\end{equation}

where $\boldsymbol{W}$ is the pretrained weights of the FFN layer, which is shared by $\{E_{k}\}_{k=1}^{K}$, $R(\cdot)$ denotes the Top-K router we employed to select specific LoRA experts for different tokens and tasks and  $E_{k}(\cdot)$ represents $k$-th LoRA experts in the \model{} module. The role of \model{} is to replace the FFN layer of the dense models in Equation~\ref{mixloraffn}, and its key concept is to select different experts by the router for each token, where each expert is composed of a different LoRA and origin FFN layer~(Equation~\ref{mixlora1}). \looseness=-1  


\paragraph{Top-K Router.}
The Top-K router in an MoE layer determines the assignment of each token to the most suitable experts~\cite{lepikhingshard}. The router is a linear layer that computes the probability of the input token $\mathbf{x}_i$ being routed to each expert: $\boldsymbol{W_r}(s)$. Within the sparse transformer block, this router activates the most appropriate LoRA experts based on the input tokens. It leverages the softmax activation function to model a probability distribution over the experts. The router's weights $\boldsymbol{W}_{r}$ are the trainable parameters of the routing network. As shown in Figure~\ref{fig:mixlora}, a top-2 gate router is employed in our design, which chooses the best two experts from $n$ available $\{E_{k}\}_{k=1}^{K}$ for each input token $\mathbf{x}_i$: 


\begin{equation}
    R^{\ell}(\mathbf{h}_i^{\ell}) = \text{KeepTop-2}(\text{Softmax}(\boldsymbol{W}^{\ell}_{r} \cdot \mathbf{x}_i)).
\end{equation}


During inference, the top-k gate router dynamically selects the best $k$ experts for each token. Through this mechanism, the mix-experts and the router work in tandem, enabling the experts to develop varied capacities and efficiently handle diverse types of tasks.

\textbf{Experts Load Balance.} Unbalanced load of experts is a significant challenge for MoEs. This is because some experts tend to be chosen more frequently by the top-k routers \cite{fedus2022switch}. To encounter this load imbalance, we apply load balancing loss to mitigate the unbalanced load for experts when training. Inspired by Switch Transformers~\cite{zeng2024turn}, we calculate the auxiliary loss and add it to a total loss. Given $N$ experts indexed by $i = 1$ to $N$ and a batch $B$ with $T$ tokens, the auxiliary loss is computed as following:
\begin{equation}
    \mathcal{L}_{\text{aux}} = a \cdot N \cdot \sum_{i = 1}^{N}{\mathcal{F}_i \cdot \mathcal{P}_i}, 
\end{equation}
\begin{equation}
    \mathcal{F}_i = \frac{1}{T}\sum_{x \in B}{\mathbbm{1}\{\operatorname{argmax}\limits_{k}{R}({x})_k=i\}}, \mathcal{P}_i = \frac{1}{T}\sum_{x \in B}{{R}(x)_i}.
\end{equation}

Where ${R}(\cdot)$ is the top-k router, $\mathcal{F}_i$ is the fraction of tokens dispatched to expert $i$ and $\mathcal{P}_i$ is the fraction of the router probability allocated for expert $i$. The final loss is multiplied by the expert count $N$ to keep the loss constant as the number of experts varies. Additionally, we utilize $a = 10^{-2}$ as a multiplicative coefficient for auxiliary losses, which is large enough to ensure load balancing while remaining small enough not to overwhelm the primary cross-entropy objective.

\textbf{Adding LoRA with Attention Layer.} \model{} further extends its fine-tuning capabilities to encompass the attention layer. Previous studies, such as ST-MoE~\cite{zoph2022stmoe}, have suggested that fine-tuning the attention layer can significantly improve performance. To enhance the fine-tuning process with \model{}, we integrate LoRA adapters into the attention layer of the dense model~(as shown in Figure~\ref{fig:mixlora}). Experimental results demonstrate that the \model{} model, fine-tuned with $q, k, v$, and $o$ projection, consistently achieves superior average scores compared to the identical configuration trained solely with a sparse layer of mixture of LoRA experts (i.e., \model{} MoE).

\subsection{Performance Optimization of \model{}} \label{sec:optimization}

\begin{figure}
    \centering
    \includegraphics[width=0.9\linewidth]{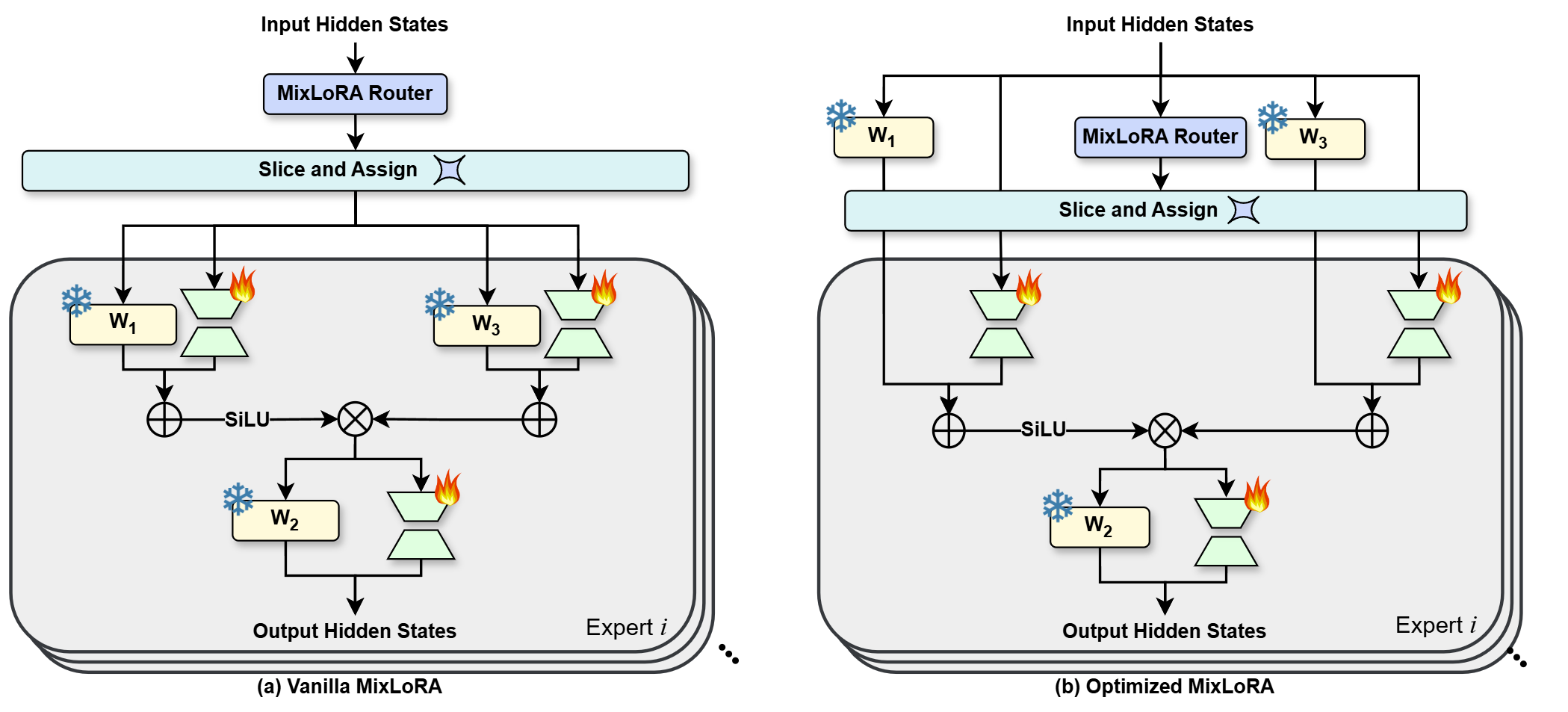}
    \caption{Comparison of the forward propagation processes: (a) the process in a vanilla \model{} MoE block; (b) the optimized process that shares computation results of $W_1$ and $W_3$ to reduce computational complexity.}
    \label{fig:perf-optim}
\end{figure}
\vspace{-2mm}

\noindent \textbf{Reducing the Computational Complexity~(I).} As describes in previous section, each expert network in \model{} includes a shared frozen FFN and several LoRAs used for storing the updated parameters of each linear projection layer of experts during fine-tuning. This setup causes the computational complexity to vary based on the $K$ settings of the router. This is because the token-level Top-K router in \model{} sends each token to $K$ experts for computation and then aggregates their residuals to produce the output. Taking LLaMA as an example, the FFN block of LLaMA consists of three linear projection weights: $W_1$, $W_2$, and $W_3$, and the forward propagation process can be expressed as $H = W_2(\text{SiLU}(W_1(x)) \cdot W_3(x))$. As shown in Figure~\ref{fig:perf-optim}~(a), each expert in \model{} has the same forward propagation process, except that every linear projection layer has a separate LoRA, and the input $x$ of each expert is pre-allocated by the \model{} router. This introduces a notable overhead when processing long sequence inputs, posing a significant challenge for the performance optimization of \model{}: \textit{How to reduce the computational complexity of \model{} while maintaining model accuracy?}



Considering we have shared the weights of the FFN block, we further share the computation results to reduce computational complexity. As shown in Figure~\ref{fig:perf-optim}~(b), rather than pre-allocating the input sequence of the \model{} block, the efficient approach first sends the input directly to $W_1$ and $W_3$ of the FFN block in parallel, then slices the output of these linear projections by the routing weights output by the \model{} router. The computational complexity of the $W_2$ projection cannot be reduced because its computation process depends on the outputs of the $W_1$ and $W_3$ projections.
This approach significantly reduces one-third of the computational complexity of the \model{} MoE block. In our experiments, the token computation latency of this approach was approximately 30\% less than that of the vanilla \model{}, while maintaining the same model performance.

\noindent \textbf{Optimizing Multi-\model{} Training and Inference~(II).} Inspired by the \textit{Multi-LoRA Optimization} proposed by m-LoRA~\cite{ye2023aspen}, we also optimized \model{} for multi-model high-throughput training and inference. Previously, we reduced the computational complexity of \model{} by eliminating duplicated calculations. When training and inferencing with two or more \model{} models, the multi-task inputs of these models are packed into a single batch to improve the training throughput.
Specifically, we first send the batched inputs into $W_1$ and $W_3$ in parallel, and then slice the outputs of these linear projections using the separate routing weights of different \model{} models.
This approach significantly reduces the memory usage of multiple \model{} models by sharing the same pre-trained model weights. In our experiments, the peak GPU memory usage of this approach was reduced by approximately 45\% per model, while maintaining the same token computation latency. We describe our optimization algorithm in detail in Appendix~\ref{appd:optimization}.

\section{Experiments} \label{sec:exp}


\subsection{Experimental Setup}

\noindent \textbf{Datasets.}
To assess \model{}, we conducted experiments on diverse commonsense reasoning datasets, including question-answer tasks (ARC~\cite{clark2018think}, OpenBookQA~\cite{mihaylov2018suit}, PIQA~\cite{bisk2019piqa}, SocialIQA~\cite{sap2019socialiqa}), a classification task (BoolQ~\cite{clark2019boolq}), a science completion task (Hellaswag~\cite{zellers2019hellaswag}), and a fill-in-the-blank task (Winogrande~\cite{sakaguchi2021winogrande}). These datasets evaluate LLMs on various challenges from scientific queries to commonsense inference. The performance of all methods was measured using the accuracy metric across all datasets. Details can be found in Appendix~\ref{appd:dataset}.

\noindent \textbf{Baselines.}
We employ commonly utilized language models: LLaMA-2 7B, LLaMA-2 13B, along with the recently introduced Gemma 2B, and LLaMA-3 8B for LoRA and DoRA, correspondingly. To maintain uniformity in parameter sizes across all PEFT methods, we instantiate LoRA and DoRA with $r = 80$ and activate $q, k, v, o$ in attention layers, as well as $w_1, w_2, w_3$ in feed-forward layers, serving as our baseline techniques. Detailed hyperparameters can be found in Appendix~\ref{appd:parameter}.

\noindent \textbf{Settings.}
In order to comprehensively evaluate the effectiveness of our method, we apply it on the basis of LoRA and DoRA, which are labeled as \textbf{\model{}} and \textbf{\doramodel{}} respectively in the experiment. Both \model{} and \doramodel{} are configured with $r = 16$, incorporating 8 experts and a top-2 router mechanism. We apply the $q, k, v, o$ parameters on the attention layers, as well as the weights $w_1, w_2, w_3$ on the feed-forward layers for the experts.

\subsection{Main Results}


\begin{table}[!t]
    \centering
    \caption{Comparison of different PEFT methods for single-task learning, using base models with different architectures and number of parameters. Reported results are accuracy scores.}
    \scalebox{0.73}{
    \begin{tabular}{lccccccccccc}
        \toprule
        \textbf{Model} & \textbf{Method} & \textbf{\# Params} & \textbf{ARC-e} & \textbf{ARC-c} & \textbf{BoolQ} & \textbf{OBQA} & \textbf{PIQA} & \textbf{SIQA} & \textbf{HellaS} & \textbf{WinoG} & \textbf{AVG.} \\ 
        \midrule
        \multirow{4}{*}{Gemma 2B} 
          & LoRA & 3.2\% & 71.9 & 43.2 & 62.1 & 71.4 & 80.9 & 71.4 & 84.4 & 46.8 & 66.5 \\ 
        ~ & DoRA & 3.2\% & 71.5 & 46.2 & 62.2 & 70.4 & 81.6 & 71.9 & 85.4 & 50.4 & 67.5 \\ 
        ~ & \textbf{MixLoRA} & 4.3\% & 76.3 & 47.4 & 65.8 & 75.8 & 81.1 & 73.6 & 89.0 & 50.4 & \textbf{69.9} \\ 
        ~ & \textbf{MixDoRA} & 4.3\% & 77.0 & 54.3 & 67.2 & 75.4 & 81.8 & 75.9 & 89.3 & 51.6 & \textbf{71.6} \\ 
        \midrule
        \multirow{4}{*}{LLaMA-2 7B}
          & LoRA & 2.9\% & 73.8  & 50.9  & 62.2  & 80.4  & 82.1  & 69.9 & 88.4 & 66.8 & 71.8 \\
        ~ & DoRA & 2.9\% & 76.5  & 59.8  & 71.7  & 80.6  & 82.7  & 74.1 & 89.6 & 67.3 & 75.3 \\ 
        ~ & \textbf{MixLoRA} & 2.9\% & 77.7 & 58.1 & 72.7 & 81.6 & 83.2 & 78.0 & 93.1 & 76.8 & \textbf{77.6} \\ 
        ~ & \textbf{MixDoRA} & 2.9\% & 77.5 & 58.2 & 72.6 & 80.9 & 82.2 & 80.4 & 90.6 & 83.4 & \textbf{78.2} \\ 
        \midrule
        \multirow{4}{*}{LLaMA-3 8B} & LoRA & 2.6\% & 89.0  & 75.7  & 67.2  & 85.0  & 80.7  & 78.3 & 74.2 & 75.3 & 78.2 \\
        ~ & DoRA & 2.6\% & 88.1  & 76.4  & 61.7  & 80.6  & 82.3  & 76.2 & 78.8 & 83.7 & 78.5 \\ 
        ~ & \textbf{MixLoRA}  & 3.0\% & 86.5 & 79.9 & 75.0 & 84.8 & 87.6 & 78.8 & 93.3 & 82.1 & \textbf{83.5}  \\ 
        ~ & \textbf{MixDoRA} & 3.0\% & 87.7 & 78.9 & 76.8 & 86.9 & 83.4 & 80.1 & 94.6 & 84.2 & \textbf{84.1} \\ 
        \midrule
        \multirow{4}{*}{LLaMA-2 13B} & LoRA & 2.4\% & 83.2  & 67.6  & 75.4  & 83.2  & 86.7  & 80.0 & 94.3 & 81.9 & 81.5 \\ 
        ~ & DoRA & 2.4\% & 83.1  & 67.7  & 75.1  & 84.5  & 87.8  & 80.1 & 94.8 & 82.4 & 81.9 \\ 
        ~ & \textbf{MixLoRA} & 2.5\% & 83.5  & 69.9  & 77.1  & 83.0  & 86.8  & 82.5 & 94.7 & 86.3 & \textbf{83.0} \\ 
        ~ & \textbf{MixDoRA} & 2.5\% & 83.7  & 68.4  & 76.9  & 83.4  & 86.9  & 83.9 & 95.2 & 86.5 & \textbf{83.1} \\
        \bottomrule
    \end{tabular}}
    
    \label{table:single_task_results}

    \vspace{4mm}
\end{table}

\noindent \textbf{Single-Task Setup.} Table \ref{table:single_task_results} compares the performance of LoRA, DoRA, \model{}, and \doramodel{} when employing these methods for fine-tuning on a single evaluation task. The results demonstrate that DoRA outperforms LoRA in most evaluations, and our methods \textbf{\model{}} and \textbf{\doramodel{}} achieve commendable performance across all evaluation metrics. However, \doramodel{} does not consistently exhibit superior performance compared to \model{}, only showing a small advantage in average accuracy.
Additionally, in most evaluations, it can be observed that the fine-tuning results of \model{} and \doramodel{} on LLaMA3 8B (83.5\%, 84.1\%) surpass the results on LLaMA2 13B (83.0\%, 83.1\%), while LoRA and DoRA leave a significant gap (8B: 78.2\%, 78.5\%; 13B: 81.5\%, 81.9\%). This indicates that our method effectively extends the model's capacity by building multiple experts on the FFN of the dense model.




\begin{table}[!t]
    \centering
        \caption{Comparison of different PEFT methods for multi-task learning, using LLaMA2 7B as the base model. Single-Task~(\textbf{ST}) setup refers to training and evaluating PEFT modules for each task, while Multi-Task~(\textbf{MT}) setup refers to training on mixed tasks, followed by separate evaluation. Reported results are accuracy scores.}
    \scalebox{0.8}{
    \begin{tabular}{ccccccccc}
        \toprule
         \textbf{PEFT Method} & \textbf{\# Params (\%)} & \textbf{ST/MT} & \textbf{ARC-e} & \textbf{ARC-c} & \textbf{BoolQ} & \textbf{OBQA} & \textbf{PIQA} & \textbf{AVG.} \\ 
        \midrule
        \multirow{2}{*}{LoRA}  & 2.9\% & ST & 73.8  & 50.9  & 62.2  & 80.4  & 82.1  & 69.9  \\ 
        ~ & 2.9\% & \textbf{MT} & 61.3  & 55.7  & 66.7  & 71.6  & 72.4  & 65.5  \\ 
        \rowcolor{Gray} 
        ~ & ~ & ~ & \textcolor{myred}{-12.5} & \textcolor{mygreen}{4.8} & \textcolor{mygreen}{4.5} & \textcolor{myred}{-8.8} & \textcolor{mygreen}{2.5} & \textcolor{myred}{-1.9} \\
        \midrule
        \multirow{2}{*}{DoRA} & 2.9\% & ST & 76.5  & 59.8  & 71.7  & 80.6  & 82.7  & 74.3  \\ 
        ~ & 2.9\% & \textbf{MT} & 64.5  & 54.1  & 65.4  & 75.8  & 71.9  & 66.3  \\ 
        \rowcolor{Gray} 
         ~ & ~ & ~ &  \textcolor{myred}{-12.0} & \textcolor{myred}{-5.7} & \textcolor{myred}{-6.3} & \textcolor{myred}{-2.8} & \textcolor{myred}{-6.9} & \textcolor{myred}{-6.7} \\ 
        \midrule
        \multirow{2}{*}{\textbf{MixLoRA}} & 2.9\% & ST & 77.7  & 58.1  & 72.7  & 81.6  & 83.2  & 74.7  \\ 
        ~ & 2.9\% & \textbf{MT} & 76.6  & 64.2  & 71.2  & 81.6  & 82.7  & 75.3  \\
        \rowcolor{Gray} 
        ~ & ~ & ~ &   \textcolor{myred}{-1.1} & \textcolor{mygreen}{6.1} & \textcolor{myred}{-1.5} & - & \textcolor{myred}{-0.5} & \textcolor{mygreen}{0.6} \\ 
        \midrule
        \multirow{2}{*}{\textbf{MixDoRA}} & 2.9\% & ST & 77.5  & 58.2  & 72.6  & 80.9  & 82.2  & 74.3  \\ 
        ~ & 2.9\% & \textbf{MT} & 76.9  & 63.4  & 71.8 & 82.2  & 80.4  & 74.9 \\ 
        \rowcolor{Gray} 
        ~ & ~ & ~ &  \textcolor{myred}{-0.6} & \textcolor{mygreen}{5.2} & \textcolor{mygreen}{0.8} & \textcolor{mygreen}{1.3} & \textcolor{myred}{-1.8} & \textcolor{mygreen}{0.6} \\ 
        \bottomrule
    \end{tabular}}
    \vspace{-2mm}
\label{table:multi_task_results_llama}
\end{table}


\noindent \textbf{Multi-Task Setup.} Table~\ref{table:multi_task_results_llama} presents the results of LoRA, DoRA, \model{}, and \doramodel{} with LLaMA2-7B in multi-task learning. In contrast to the single-task learning results shown in Table \ref{table:single_task_results}, during multi-task learning, we mixed training data from ARC, BoolQ, OBQA, and PIQA to train the model, followed by separate evaluations to investigate the generalization ability of each method. The results indicate that, compared to single-task learning, LoRA and DoRA exhibit degradation in average accuracy in multi-task learning (LoRA: -1.9\%, DoRA: -6.7\%), while \model{} and \doramodel{} maintain nearly the same average accuracy.
Additionally, we conducted multi-task learning on Gemma 2B, with results presented in Appendix~\ref{appd:ml_gemma}. \model{} and \doramodel{} maintain their superiority, showing the lowest performance degradation.
This suggests that \model{} and \doramodel{} demonstrate stronger generalization ability and mitigate the issue of knowledge forgetting in multi-task learning compared to LoRA and DoRA.


\begin{figure}[!t]
\begin{center}
\begin{minipage}{0.32\textwidth}
\includegraphics[width=\linewidth]{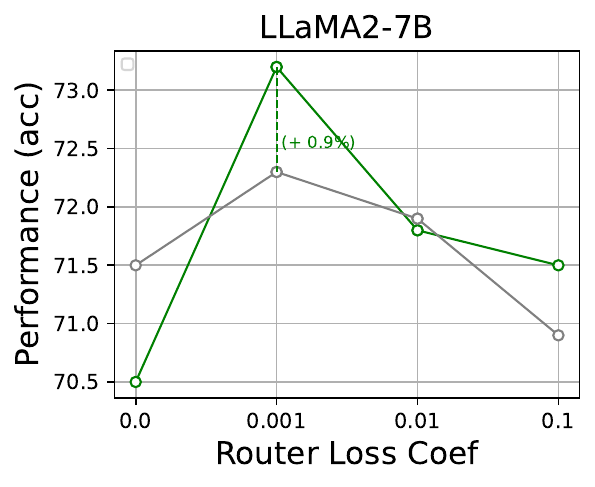}
\caption*{(a)}
\end{minipage}
\begin{minipage}{0.32\textwidth}
\includegraphics[width=\linewidth]{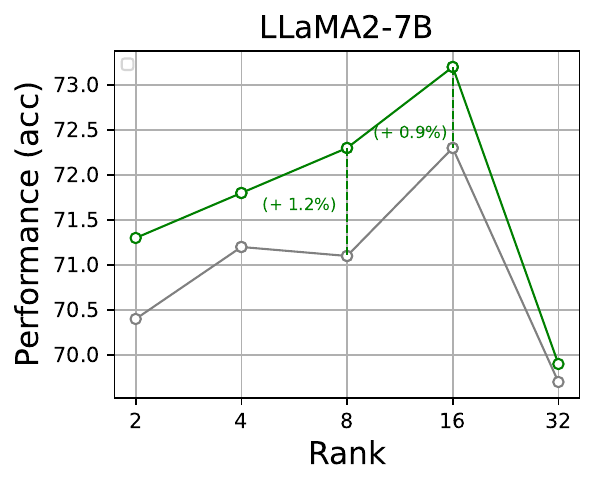}
\caption*{(b)}
\end{minipage}
\begin{minipage}{0.32\textwidth}
\includegraphics[width=\linewidth]{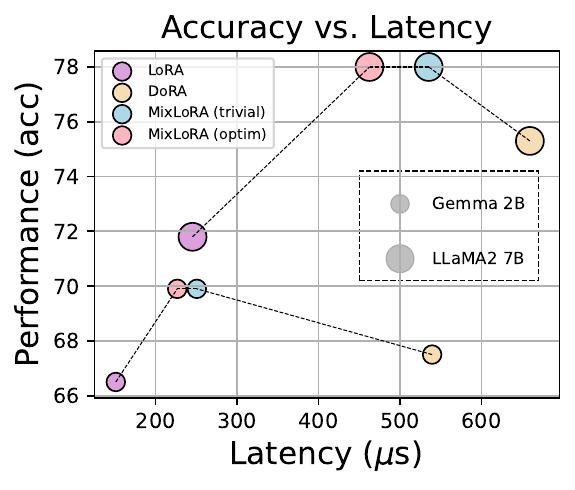}
\caption*{(c)}
\end{minipage}
\caption{Ablation studies on router loss coefficient~(a) and rank~(b) on LLaMA2 7B. (c) \model{} outperforms LoRA and DoRA without introducing significant latency on diverse commonsense tasks.}
\label{fig:rank_loss_efficiency}
\end{center}
\vskip -0.3in
\end{figure}

\begin{figure}[h]
    \centering
    \begin{minipage}{0.5\textwidth}
        \centering
        \includegraphics[width=1\textwidth]{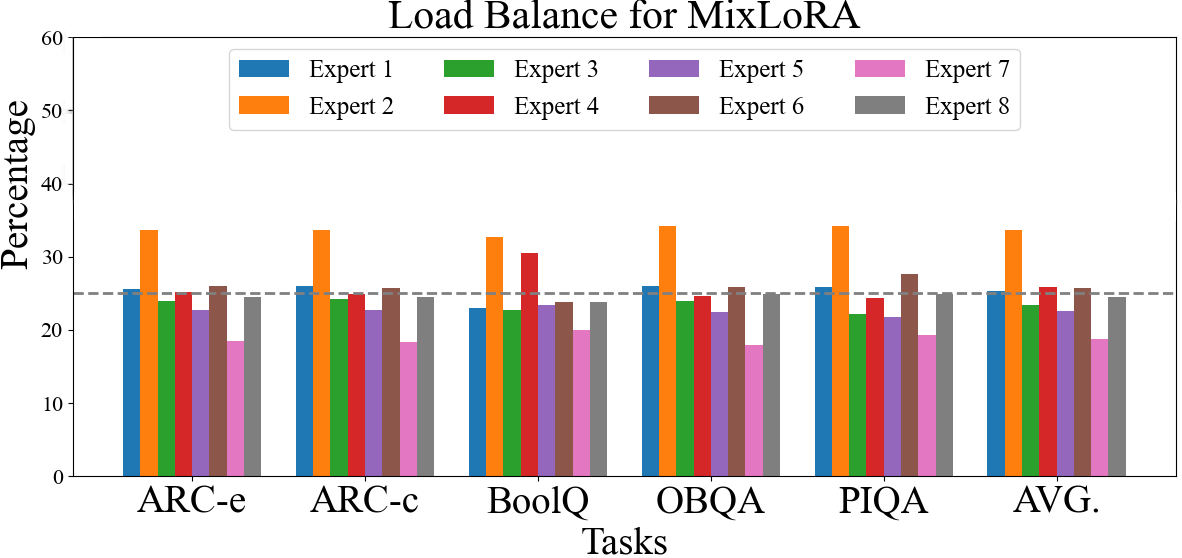} 
        \caption*{(a)}
    \end{minipage}\hfill
    \begin{minipage}{0.5\textwidth}
        \centering
        \includegraphics[width=1\textwidth]{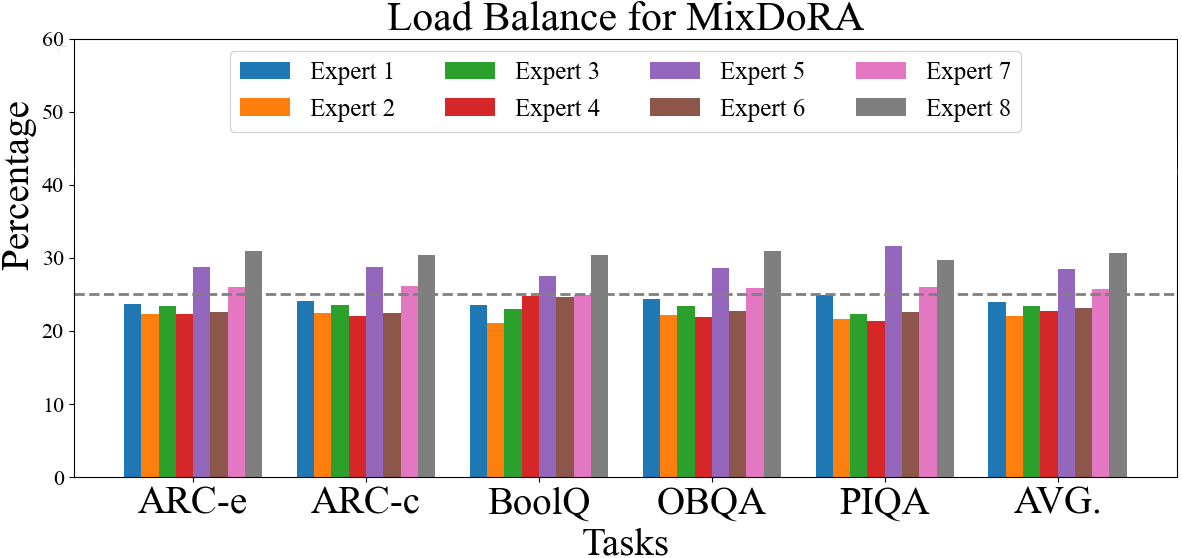} 
        \caption*{(b)}
    \end{minipage}
    \caption{Distribution of expert loadings. The average workload of the 8 experts in \model{} (a) and \doramodel{} (b) during the evaluation of multi-task learning is depicted in the figure. The average standard deviation of \model{} is smaller than that of \doramodel{} (0.0223 < 0.0328). However, both standard deviations are small enough, indicating that the workload of these experts is balanced.}
    \label{fig:expert_load}
\end{figure}

\subsection{Ablation Study}

\noindent \textbf{Analysis of Auxiliary Loss.} We investigated the influence of different router loss coefficients on model performance with ARC, OpenBookQA, and BoolQ using LLaMA2-7B. As shown in Figure~\ref{fig:rank_loss_efficiency}~(a), results indicate that with the coefficient of 1e-3, \model{} achieves the highest average accuracy. Disabling router loss or using a higher coefficient results in lower average accuracy. This suggests that a reasonable router loss coefficient can help address the imbalance problem of experts, while a higher coefficient can impede model convergence during fine-tuning. Furthermore, \model{} imposes higher requirements on the rationality of the router loss coefficient compared to \doramodel{}, indicating that MixDoRA is less sensitive to hyperparameter settings than \model{}. For example, when disabling router loss (coefficient = 0), \model{} lags behind \doramodel{} by 1\% in average accuracy, while surpassing \doramodel{} by 0.9\% in average accuracy when the coefficient is set to 1e-3. \looseness=-1

To confirm the effectiveness of auxiliary loss, we also present the distribution of experts across diverse tasks for \model{} and \doramodel{} in Figure~\ref{fig:expert_load}. Intuitively, we can observe that the expert loads in all tasks are balanced, indicating that these experts are allocated to different tasks evenly. Specifically, both \model{} and \doramodel{} achieves low average standard deviation (0.0223 and 0.0328). This verifies that the use of auxiliary load balance loss mitigates the imbalance problem of experts.

\noindent \textbf{Analysis of Model Rank.} We investigated the influence of different ranks on model performance with ARC, OpenBookQA, and BoolQ using LLaMA2-7B. As shown in Figure \ref{fig:rank_loss_efficiency}~(b), both \model{} and \doramodel{} consistently perform well from rank 2 to rank 16. However, the average accuracy of them both drops when rank is 32 due to convergence difficulties. Moreover, \doramodel{} slightly lags behind \model{} at all ranks. This is because our method introduces more fine-tuning diversity through the MoE structure, making DoRA's weight decomposition method less effective when applied to our method (\doramodel{}). For example, in our previous multi-task experiments on Tabel~\ref{table:multi_task_results_llama}, DoRA often showed more degradation compared to \model{} and even LoRA.

\noindent \textbf{Analysis of Computation Efficiency.} To measure the computational overhead between LoRA, DoRA, and \model{}, we collected the token computation latency and peak GPU memory usage from PEFT for LoRA and DoRA, and from m-LoRA for \model{} during training and inference separately. The trainable parameters for all methods were strictly controlled to be equal both on Gemma 2B and LLaMA2 7B. Figure~\ref{fig:rank_loss_efficiency}~(c) shows the overall comparison of model accuracy and token computation latency between different model sizes (2B and 7B). Our method, \model{}, achieves the best accuracy in both Gemma 2B and LLaMA2 7B, while the token computation latency falls between LoRA and DoRA. The optimized \model{} (in Section~\ref{sec:optimization}) shows significant improvement while maintaining the same model accuracy, especially with the larger 7B model. Details can be found in Appendix~\ref{appd:efficiency}.

\clearpage

\section{Conclusion} \label{sec:conclusion}

In this paper, we introduce \model{}, a parameter-efficient MoE method using multiple LoRA based experts and a frozen shared FFN block. Unlike traditional LoRA-MoE approaches, \model{} fused multiple LoRAs with the shared FFN layer and employs them to store updated parameters for each expert during fine-tuning, aligning it more with pre-trained MoE models. It also employs self-attention LoRA adapters and an auxiliary load balance loss to improve performance and address router imbalance. Furthermore, we design a high-performance framework to optimize the computation process of multiple LoRA based experts in \model{} both for training and inference. As a result, this framework reduces the computational complexity of \model{} by 30\%, and saves about 40\% GPU memory usage when training or inferencing multiple \model{} models. Evaluation shows \model{} outperforms baselines in single-task and multi-task scenarios. For single-task learning, \model{} achieves an average accuracy improvement of \textbf{5.8\%} on LLaMA-2 7B compared to LoRA and \textbf{2.3\%} compared to DoRA. In multi-task learning, \model{} significantly surpasses LoRA by \textbf{9.8\%} and DoRA by \textbf{9\%} in accuracy.

\clearpage
\bibliographystyle{unsrt}  
\bibliography{references}  

\begin{thebibliography}{10}

\bibitem{Brown2020LanguageMA}
Tom~B. Brown, Benjamin Mann, Nick Ryder, Melanie Subbiah, Jared Kaplan, Prafulla Dhariwal, Arvind Neelakantan, Pranav Shyam, Girish Sastry, Amanda Askell, Sandhini Agarwal, Ariel Herbert-Voss, Gretchen Krueger, T.~J. Henighan, Rewon Child, Aditya Ramesh, Daniel~M. Ziegler, Jeff Wu, Clemens Winter, Christopher Hesse, Mark Chen, Eric Sigler, Mateusz Litwin, Scott Gray, Benjamin Chess, Jack Clark, Christopher Berner, Sam McCandlish, Alec Radford, Ilya Sutskever, and Dario Amodei.
\newblock Language models are few-shot learners.
\newblock {\em ArXiv}, abs/2005.14165, 2020.

\bibitem{Chowdhery2022PaLMSL}
Aakanksha Chowdhery, Sharan Narang, Jacob Devlin, Maarten Bosma, Gaurav Mishra, Adam Roberts, Paul Barham, Hyung~Won Chung, Charles Sutton, Sebastian Gehrmann, Parker Schuh, Kensen Shi, Sasha Tsvyashchenko, Joshua Maynez, Abhishek Rao, Parker Barnes, Yi~Tay, Noam~M. Shazeer, Vinodkumar Prabhakaran, Emily Reif, Nan Du, Benton~C. Hutchinson, Reiner Pope, James Bradbury, Jacob Austin, Michael Isard, Guy Gur-Ari, Pengcheng Yin, Toju Duke, Anselm Levskaya, Sanjay Ghemawat, Sunipa Dev, Henryk Michalewski, Xavier Garc{\'i}a, Vedant Misra, Kevin Robinson, Liam Fedus, Denny Zhou, Daphne Ippolito, David Luan, Hyeontaek Lim, Barret Zoph, Alexander Spiridonov, Ryan Sepassi, David Dohan, Shivani Agrawal, Mark Omernick, Andrew~M. Dai, Thanumalayan~Sankaranarayana Pillai, Marie Pellat, Aitor Lewkowycz, Erica Moreira, Rewon Child, Oleksandr Polozov, Katherine Lee, Zongwei Zhou, Xuezhi Wang, Brennan Saeta, Mark D{\'i}az, Orhan Firat, Michele Catasta, Jason Wei, Kathleen~S. Meier-Hellstern, Douglas Eck, Jeff Dean, Slav Petrov,
  and Noah Fiedel.
\newblock Palm: Scaling language modeling with pathways.
\newblock {\em J. Mach. Learn. Res.}, 24, 2022.

\bibitem{Hoffmann2022TrainingCL}
Jordan Hoffmann, Sebastian Borgeaud, Arthur Mensch, Elena Buchatskaya, Trevor Cai, Eliza Rutherford, Diego de~Las~Casas, Lisa~Anne Hendricks, Johannes Welbl, Aidan Clark, Tom Hennigan, Eric Noland, Katie Millican, George van~den Driessche, Bogdan Damoc, Aurelia Guy, Simon Osindero, Karen Simonyan, Erich Elsen, Jack~W. Rae, Oriol Vinyals, and L.~Sifre.
\newblock Training compute-optimal large language models.
\newblock {\em ArXiv}, abs/2203.15556, 2022.

\bibitem{Touvron2023LLaMAOA}
Hugo Touvron, Thibaut Lavril, Gautier Izacard, Xavier Martinet, Marie-Anne Lachaux, Timoth{\'e}e Lacroix, Baptiste Rozi{\`e}re, Naman Goyal, Eric Hambro, Faisal Azhar, Aurelien Rodriguez, Armand Joulin, Edouard Grave, and Guillaume Lample.
\newblock Llama: Open and efficient foundation language models.
\newblock {\em ArXiv}, abs/2302.13971, 2023.

\bibitem{Touvron2023Llama2O}
Hugo Touvron, Louis Martin, Kevin~R. Stone, Peter Albert, Amjad Almahairi, Yasmine Babaei, Nikolay Bashlykov, Soumya Batra, Prajjwal Bhargava, Shruti Bhosale, Daniel~M. Bikel, Lukas Blecher, Cristian~Cant{\'o}n Ferrer, Moya Chen, Guillem Cucurull, David Esiobu, Jude Fernandes, Jeremy Fu, Wenyin Fu, Brian Fuller, Cynthia Gao, Vedanuj Goswami, Naman Goyal, Anthony~S. Hartshorn, Saghar Hosseini, Rui Hou, Hakan Inan, Marcin Kardas, Viktor Kerkez, Madian Khabsa, Isabel~M. Kloumann, A.~V. Korenev, Punit~Singh Koura, Marie-Anne Lachaux, Thibaut Lavril, Jenya Lee, Diana Liskovich, Yinghai Lu, Yuning Mao, Xavier Martinet, Todor Mihaylov, Pushkar Mishra, Igor Molybog, Yixin Nie, Andrew Poulton, Jeremy Reizenstein, Rashi Rungta, Kalyan Saladi, Alan Schelten, Ruan Silva, Eric~Michael Smith, R.~Subramanian, Xia Tan, Binh Tang, Ross Taylor, Adina Williams, Jian~Xiang Kuan, Puxin Xu, Zhengxu Yan, Iliyan Zarov, Yuchen Zhang, Angela Fan, Melanie Kambadur, Sharan Narang, Aurelien Rodriguez, Robert Stojnic, Sergey Edunov, and
  Thomas Scialom.
\newblock Llama 2: Open foundation and fine-tuned chat models.
\newblock {\em ArXiv}, abs/2307.09288, 2023.

\bibitem{Chung2022ScalingIL}
Hyung~Won Chung, Le~Hou, S.~Longpre, Barret Zoph, Yi~Tay, William Fedus, Eric Li, Xuezhi Wang, Mostafa Dehghani, Siddhartha Brahma, Albert Webson, Shixiang~Shane Gu, Zhuyun Dai, Mirac Suzgun, Xinyun Chen, Aakanksha Chowdhery, Dasha Valter, Sharan Narang, Gaurav Mishra, Adams~Wei Yu, Vincent Zhao, Yanping Huang, Andrew~M. Dai, Hongkun Yu, Slav Petrov, Ed~Huai hsin Chi, Jeff Dean, Jacob Devlin, Adam Roberts, Denny Zhou, Quoc~V. Le, and Jason Wei.
\newblock Scaling instruction-finetuned language models.
\newblock {\em ArXiv}, abs/2210.11416, 2022.

\bibitem{Iyer2022OPTIMLSL}
Srinivas Iyer, Xi~Victoria Lin, Ramakanth Pasunuru, Todor Mihaylov, Daniel Simig, Ping Yu, Kurt Shuster, Tianlu Wang, Qing Liu, Punit~Singh Koura, Xian Li, Brian O'Horo, Gabriel Pereyra, Jeff Wang, Christopher Dewan, Asli Celikyilmaz, Luke Zettlemoyer, and Veselin Stoyanov.
\newblock Opt-iml: Scaling language model instruction meta learning through the lens of generalization.
\newblock {\em ArXiv}, abs/2212.12017, 2022.

\bibitem{zheng2023judging}
Lianmin Zheng, Wei-Lin Chiang, Ying Sheng, Siyuan Zhuang, Zhanghao Wu, Yonghao Zhuang, Zi~Lin, Zhuohan Li, Dacheng Li, Eric Xing, et~al.
\newblock Judging llm-as-a-judge with mt-bench and chatbot arena.
\newblock {\em NeurIPS}, 36, 2024.

\bibitem{wei2022emergent}
Jason Wei, Yi~Tay, Rishi Bommasani, Colin Raffel, Barret Zoph, Sebastian Borgeaud, Dani Yogatama, Maarten Bosma, Denny Zhou, Donald Metzler, et~al.
\newblock Emergent abilities of large language models.
\newblock {\em arXiv preprint arXiv: 2206.07682}, 2022.

\bibitem{houlsby2019parameter}
Neil Houlsby, Andrei Giurgiu, Stanislaw Jastrzebski, Bruna Morrone, Quentin De~Laroussilhe, Andrea Gesmundo, Mona Attariyan, and Sylvain Gelly.
\newblock Parameter-efficient transfer learning for nlp.
\newblock In {\em ICML}, pages 2790--2799. PMLR, 2019.

\bibitem{Li2021PrefixTuningOC}
Xiang~Lisa Li and Percy Liang.
\newblock Prefix-tuning: Optimizing continuous prompts for generation.
\newblock {\em ACL}, 2021.

\bibitem{Lester2021ThePO}
Brian Lester, Rami Al-Rfou, and Noah Constant.
\newblock The power of scale for parameter-efficient prompt tuning.
\newblock In {\em EMNLP}, 2021.

\bibitem{BenZaken2021BitFitSP}
Elad Ben-Zaken, Shauli Ravfogel, and Yoav Goldberg.
\newblock Bitfit: Simple parameter-efficient fine-tuning for transformer-based masked language-models.
\newblock {\em ArXiv}, 2021.

\bibitem{Liu2022FewShotPF}
Haokun Liu, Derek Tam, Mohammed Muqeeth, Jay Mohta, Tenghao Huang, Mohit Bansal, and Colin Raffel.
\newblock Few-shot parameter-efficient fine-tuning is better and cheaper than in-context learning.
\newblock {\em ArXiv}, 2022.

\bibitem{Hu2021LoRALA}
J.~Edward Hu, Yelong Shen, Phillip Wallis, Zeyuan Allen-Zhu, Yuanzhi Li, Shean Wang, and Weizhu Chen.
\newblock Lora: Low-rank adaptation of large language models.
\newblock {\em ArXiv}, abs/2106.09685, 2021.

\bibitem{sun2022recent}
Zehua Sun, Huanqi Yang, Kai Liu, Zhimeng Yin, Zhenjiang Li, and Weitao Xu.
\newblock Recent advances in lora: A comprehensive survey.
\newblock {\em ACM Transactions on Sensor Networks}, 2022.

\bibitem{sundaram2019survey}
Jothi Prasanna~Shanmuga Sundaram, Wan Du, and Zhiwei Zhao.
\newblock A survey on lora networking: Research problems, current solutions, and open issues.
\newblock {\em IEEE Communications Surveys \& Tutorials}, 2019.

\bibitem{hayou2024lora+}
Soufiane Hayou, Nikhil Ghosh, and Bin Yu.
\newblock Lora+: Efficient low rank adaptation of large models.
\newblock {\em arXiv preprint arXiv: 2402.12354}, 2024.

\bibitem{liu2024dora}
Shih-Yang Liu, Chien-Yi Wang, Hongxu Yin, Pavlo Molchanov, Yu-Chiang~Frank Wang, Kwang-Ting Cheng, and Min-Hung Chen.
\newblock Dora: Weight-decomposed low-rank adaptation.
\newblock {\em arXiv preprint arXiv: 2402.09353}, 2024.

\bibitem{feng2024mixtureofloras}
Wenfeng Feng, Chuzhan Hao, Yuewei Zhang, Yu~Han, and Hao Wang.
\newblock Mixture-of-loras: An efficient multitask tuning for large language models.
\newblock {\em arXiv preprint arXiv: 2403.03432}, 2024.

\bibitem{huang2024lorahub}
Chengsong Huang, Qian Liu, Bill~Yuchen Lin, Tianyu Pang, Chao Du, and Min Lin.
\newblock Lorahub: Efficient cross-task generalization via dynamic lora composition.
\newblock {\em arXiv preprint arXiv: 2307.13269}, 2023.

\bibitem{pfeiffer2020adapterfusion}
Jonas Pfeiffer, Aishwarya Kamath, Andreas R{\"u}ckl{\'e}, Kyunghyun Cho, and Iryna Gurevych.
\newblock Adapterfusion: Non-destructive task composition for transfer learning.
\newblock {\em arXiv preprint arXiv: 2005.00247}, 2020.

\bibitem{yang2024moral}
Shu Yang, Muhammad~Asif Ali, Cheng-Long Wang, Lijie Hu, and Di~Wang.
\newblock Moral: Moe augmented lora for llms' lifelong learning.
\newblock {\em arXiv preprint arXiv: 2402.11260}, 2024.

\bibitem{luo2024moelora}
Tongxu Luo, Jiahe Lei, Fangyu Lei, Weihao Liu, Shizhu He, Jun Zhao, and Kang Liu.
\newblock Moelora: Contrastive learning guided mixture of experts on parameter-efficient fine-tuning for large language models.
\newblock {\em arXiv preprint arXiv: 2402.12851}, 2024.

\bibitem{wu2024parameterefficient}
Haoyuan Wu, Haisheng Zheng, and Bei Yu.
\newblock Parameter-efficient sparsity crafting from dense to mixture-of-experts for instruction tuning on general tasks.
\newblock {\em arXiv preprint arXiv: 2401.02731}, 2024.

\bibitem{dou2024loramoe}
Shihan Dou, Enyu Zhou, Yan Liu, Songyang Gao, Jun Zhao, Wei Shen, Yuhao Zhou, Zhiheng Xi, Xiao Wang, Xiaoran Fan, Shiliang Pu, Jiang Zhu, Rui Zheng, Tao Gui, Qi~Zhang, and Xuanjing Huang.
\newblock Loramoe: Alleviate world knowledge forgetting in large language models via moe-style plugin.
\newblock {\em arXiv}, 2024.

\bibitem{gou2024mixture}
Yunhao Gou, Zhili Liu, Kai Chen, Lanqing Hong, Hang Xu, Aoxue Li, Dit-Yan Yeung, James~T Kwok, and Yu~Zhang.
\newblock Mixture of cluster-conditional lora experts for vision-language instruction tuning.
\newblock {\em arXiv preprint arXiv: 2312.12379}, 2023.

\bibitem{Liu2023MOELoRAAM}
Qidong Liu, Xian Wu, Xiangyu Zhao, Yuanshao Zhu, Derong Xu, Feng Tian, and Yefeng Zheng.
\newblock Moelora: An moe-based parameter efficient fine-tuning method for multi-task medical applications.
\newblock {\em ArXiv: 2310.18339}, 2023.

\bibitem{gao2024higher}
Chongyang Gao, Kezhen Chen, Jinmeng Rao, Baochen Sun, Ruibo Liu, Daiyi Peng, Yawen Zhang, Xiaoyuan Guo, Jie Yang, and VS~Subrahmanian.
\newblock Higher layers need more lora experts.
\newblock {\em arXiv preprint arXiv: 2402.08562}, 2024.

\bibitem{jiang2024mixtral}
Albert~Q Jiang, Alexandre Sablayrolles, Antoine Roux, Arthur Mensch, Blanche Savary, Chris Bamford, Devendra~Singh Chaplot, Diego de~las Casas, Emma~Bou Hanna, Florian Bressand, et~al.
\newblock Mixtral of experts.
\newblock {\em arXiv preprint arXiv: 2401.04088}, 2024.

\bibitem{vaswani2023attention}
Ashish Vaswani, Noam Shazeer, Niki Parmar, Jakob Uszkoreit, Llion Jones, Aidan~N Gomez, {\L}ukasz Kaiser, and Illia Polosukhin.
\newblock Attention is all you need.
\newblock {\em NeurIPS}, 2017.

\bibitem{zoph2022stmoe}
Barret Zoph, Irwan Bello, Sameer Kumar, Nan Du, Yanping Huang, Jeff Dean, Noam Shazeer, and William Fedus.
\newblock St-moe: Designing stable and transferable sparse expert models.
\newblock {\em arXiv preprint arXiv: 2202.08906}, 2022.

\bibitem{ye2023aspen}
Zhengmao Ye, Dengchun Li, Jingqi Tian, Tingfeng Lan, Jie Zuo, Lei Duan, Hui Lu, Yexi Jiang, Jian Sha, Ke~Zhang, and Mingjie Tang.
\newblock Aspen: High-throughput lora fine-tuning of large language models with a single gpu.
\newblock {\em arXiv: 2312.02515}, 2023.

\bibitem{fedus2022switch}
William Fedus, Barret Zoph, and Noam Shazeer.
\newblock Switch transformers: Scaling to trillion parameter models with simple and efficient sparsity.
\newblock {\em JMLR}, 2022.

\bibitem{clark2018think}
Peter Clark, Isaac Cowhey, Oren Etzioni, Tushar Khot, Ashish Sabharwal, Carissa Schoenick, and Oyvind Tafjord.
\newblock Think you have solved question answering? try arc, the ai2 reasoning challenge.
\newblock {\em arXiv preprint arXiv: 1803.05457}, 2018.

\bibitem{clark2019boolq}
Christopher Clark, Kenton Lee, Ming-Wei Chang, Tom Kwiatkowski, Michael Collins, and Kristina Toutanova.
\newblock Boolq: Exploring the surprising difficulty of natural yes/no questions.
\newblock {\em arXiv preprint arXiv: 1905.10044}, 2019.

\bibitem{mihaylov2018suit}
Todor Mihaylov, Peter Clark, Tushar Khot, and Ashish Sabharwal.
\newblock Can a suit of armor conduct electricity? a new dataset for open book question answering.
\newblock {\em arXiv preprint arXiv: 1809.02789}, 2018.

\bibitem{bisk2020piqa}
Yonatan Bisk, Rowan Zellers, Jianfeng Gao, Yejin Choi, et~al.
\newblock Piqa: Reasoning about physical commonsense in natural language.
\newblock {\em AAAI}, 2020.

\bibitem{sap2019socialiqa}
Maarten Sap, Hannah Rashkin, Derek Chen, Ronan LeBras, and Yejin Choi.
\newblock Socialiqa: Commonsense reasoning about social interactions.
\newblock {\em arXiv preprint arXiv: 1904.09728}, 2019.

\bibitem{zellers2019hellaswag}
Rowan Zellers, Ari Holtzman, Yonatan Bisk, Ali Farhadi, and Yejin Choi.
\newblock Hellaswag: Can a machine really finish your sentence?
\newblock {\em arXiv preprint arXiv: 1905.07830}, 2019.

\bibitem{sakaguchi2021winogrande}
Keisuke Sakaguchi, Ronan~Le Bras, Chandra Bhagavatula, and Yejin Choi.
\newblock Winogrande: An adversarial winograd schema challenge at scale.
\newblock {\em Communications of the ACM}, 2021.

\bibitem{chatgpt}
OpenAI.
\newblock Chatgpt: Optimizing language models for dialogue., 2022.

\bibitem{gpt4}
Josh Achiam, Steven Adler, Sandhini Agarwal, Lama Ahmad, Ilge Akkaya, Florencia~Leoni Aleman, Diogo Almeida, Janko Altenschmidt, Sam Altman, Shyamal Anadkat, et~al.
\newblock Gpt-4 technical report.
\newblock {\em arXiv preprint arXiv: 2303.08774}, 2023.

\bibitem{peft}
Sourab Mangrulkar, Sylvain Gugger, Lysandre Debut, Younes Belkada, Sayak Paul, and Benjamin Bossan.
\newblock Peft: State-of-the-art parameter-efficient fine-tuning methods.
\newblock \url{https: //github.com/huggingface/peft}, 2022.

\bibitem{Liu2023LLMQATDQ}
Zechun Liu, Barlas Oğuz, Changsheng Zhao, Ernie Chang, Pierre Stock, Yashar Mehdad, Yangyang Shi, Raghuraman Krishnamoorthi, and Vikas Chandra.
\newblock Llm-qat: Data-free quantization aware training for large language models.
\newblock {\em ArXiv}, abs/2305.17888, 2023.

\bibitem{Xiao2023OffsiteTuningTL}
Guangxuan Xiao, Ji~Lin, and Song Han.
\newblock Offsite-tuning: Transfer learning without full model.
\newblock {\em ArXiv}, abs/2302.04870, 2023.

\bibitem{Frantar2022GPTQAP}
Elias Frantar, Saleh Ashkboos, Torsten Hoefler, and Dan Alistarh.
\newblock Gptq: Accurate post-training quantization for generative pre-trained transformers.
\newblock {\em ArXiv}, abs/2210.17323, 2022.

\bibitem{Xiao2022SmoothQuantAA}
Guangxuan Xiao, Ji~Lin, Mickael Seznec, Julien Demouth, and Song Han.
\newblock Smoothquant: Accurate and efficient post-training quantization for large language models.
\newblock {\em ArXiv}, abs/2211.10438, 2022.

\bibitem{Frantar2023SparseGPTML}
Elias Frantar and Dan Alistarh.
\newblock Sparsegpt: Massive language models can be accurately pruned in one-shot.
\newblock {\em ArXiv}, abs/2301.00774, 2023.

\bibitem{Ma2023LLMPrunerOT}
Xinyin Ma, Gongfan Fang, and Xinchao Wang.
\newblock Llm-pruner: On the structural pruning of large language models.
\newblock {\em ArXiv}, abs/2305.11627, 2023.

\bibitem{kopiczko2024vera}
Dawid~Jan Kopiczko, Tijmen Blankevoort, and Yuki~Markus Asano.
\newblock Vera: Vector-based random matrix adaptation.
\newblock {\em arXiv preprint arXiv: 2310.11454}, 2023.

\bibitem{hyeonwoo2023fedpara}
Nam Hyeon-Woo, Moon Ye-Bin, and Tae-Hyun Oh.
\newblock Fedpara: Low-rank hadamard product for communication-efficient federated learning.
\newblock {\em arXiv preprint arXiv: 2108.06098}, 2021.

\bibitem{renduchintala2023tiedlora}
Adithya Renduchintala, Tugrul Konuk, and Oleksii Kuchaiev.
\newblock Tied-lora: Enhacing parameter efficiency of lora with weight tying.
\newblock {\em arXiv preprint arXiv: 2311.09578}, 2023.

\bibitem{zhang2023adalora}
Qingru Zhang, Minshuo Chen, Alexander Bukharin, Pengcheng He, Yu~Cheng, Weizhu Chen, and Tuo Zhao.
\newblock Adaptive budget allocation for parameter-efficient fine-tuning.
\newblock In {\em ICLR}, 2023.

\bibitem{Jacobs1991AdaptiveMO}
Robert~A. Jacobs, Michael~I. Jordan, Steven~J. Nowlan, and Geoffrey~E. Hinton.
\newblock Adaptive mixtures of local experts.
\newblock {\em Neural Computation}, 1991.

\bibitem{chen2024llavamole}
Shaoxiang Chen, Zequn Jie, and Lin Ma.
\newblock Llava-mole: Sparse mixture of lora experts for mitigating data conflicts in instruction finetuning mllms.
\newblock {\em arXiv preprint arXiv: 2401.16160}, 2024.

\bibitem{lepikhingshard}
Dmitry Lepikhin, HyoukJoong Lee, Yuanzhong Xu, Dehao Chen, Orhan Firat, Yanping Huang, Maxim Krikun, Noam Shazeer, and Zhifeng Chen.
\newblock Gshard: Scaling giant models with conditional computation and automatic sharding.
\newblock In {\em ICLR}, 2020.

\bibitem{wang2019learning}
Qiang Wang, Bei Li, Tong Xiao, Jingbo Zhu, Changliang Li, Derek~F Wong, and Lidia~S Chao.
\newblock Learning deep transformer models for machine translation.
\newblock {\em arXiv preprint arXiv: 1906.01787}, 2019.

\bibitem{baevski2018adaptive}
Alexei Baevski and Michael Auli.
\newblock Adaptive input representations for neural language modeling.
\newblock {\em arXiv preprint arXiv: 1809.10853}, 2018.

\bibitem{zeng2024turn}
Zhiyuan Zeng, Qipeng Guo, Zhaoye Fei, Zhangyue Yin, Yunhua Zhou, Linyang Li, Tianxiang Sun, Hang Yan, Dahua Lin, and Xipeng Qiu.
\newblock Turn waste into worth: Rectifying top-$ k $ router of moe.
\newblock {\em arXiv preprint arXiv: 2402.12399}, 2024.

\bibitem{bisk2019piqa}
Yonatan Bisk, Rowan Zellers, Ronan~Le Bras, Jianfeng Gao, and Yejin Choi.
\newblock Piqa: Reasoning about physical commonsense in natural language.
\newblock {\em arXiv: 1911.11641}, 2019.

\end{thebibliography}

\clearpage
\appendix
\section{Appendix}

\subsection{Hyperparameters and Implementation Details} \label{appd:parameter}
\begin{table}[h]
\caption{Hyperparameter configurations of LoRA/DoRA and MixLoRA/MixDoRA for fine-tuning Gemma-2B, LLaMA2-7B/13B, and LLaMA3-8B on the commonsense reasoning tasks. }
\label{table:hyperparameters}
\centering
\begin{tabular}{l|cc}
\toprule
\textbf{Hyperparameters}      & \textbf{LoRA/DoRA} & \textbf{MixLoRA/MixDoRA}   \\
\midrule
Cutoff Length &\multicolumn{2}{c}{512}\\
Learning Rate &\multicolumn{2}{c}{2e-4}\\
Optimizer     &\multicolumn{2}{c}{AdamW}\\
Batch size    &\multicolumn{2}{c}{16}\\
Accumulation Steps  &\multicolumn{2}{c}{8}\\
Dropout       &\multicolumn{2}{c}{0.05}\\
\# Epochs        &\multicolumn{2}{c}{2}\\
Where &\multicolumn{2}{c}{Q, K, V, O, Up, Down, Gate}\\
LoRA Rank $r$        &80  &16\\
LoRA Alpha $\alpha$       &160 &32\\
\# Experts        & - &8\\
Top-K         & -  &2\\
\bottomrule

\end{tabular}
\end{table}

All experiments are conducted with GPUs having 24GB memory (RTX 3090, RTX A5000, RTX 4090) for 7B models, GPUs having 48GB memory (RTX A6000) for 8B and 13B models, and setup with Python 3.10 and Ubuntu 22.04 on x86-64 CPUs.

\subsection{Datasets} \label{appd:dataset}
Table 4 presents detailed information about the datasets used in our experiments, including their task names, respective domains, the number of training and test sets, task types.
\begin{table}[ht]
\centering
\caption{Description of Datasets used in experiments.}
\begin{tabular}{l|lccr}
\toprule
\textbf{Task Name} & \textbf{Domain} & \textbf{\# Train} & \textbf{\# Test} & \textbf{Task Type}\\
\midrule
BoolQ & Wikipedia & 9,427 & 3,270 & Text Classification \\
ARC-E & Natural Science & 2,250 & 2,380 & Question Answering \\
ARC-C & Natural Science & 1,120 & 1,170 & Question Answering \\
OpenBookQA & Science Facts & 4,957 & 500 & Question Answering \\
PIQA & Physical Interaction & 16,100 & 1,840 & Question Answering \\
SIQA & Social Interaction & 33,410 & 1,954 & Question Answering \\
HellaSwag & Video Caption & 39,905 & 10,042 & Sentence Completion \\
WinoGrande & Winograd Schemas & 9,248 & 1,267 & Fill in the Blank \\
\bottomrule
\end{tabular}
\label{tab:dataset_description}
\end{table}

All datasets are downloaded from \href{https://huggingface.co}{HuggingFace} using the \textsc{Datasets} library in Python.

\newpage
\subsection{Multi-task Learning Evaluation Result using Gemma-2B} \label{appd:ml_gemma}
\begin{table}[h]
    \centering
        \caption{Comparision of different peft methods for multi-task learning on various tasks, using Gemma 2B as the base model. Single-Task~(\textbf{ST}) setup refers to training and evaluating PEFT modules for each task, while Multi-Task~(\textbf{MT}) setup refers to training on mixed tasks, followed by separate evaluation. Reported results are accuracy scores.}
    \scalebox{0.8}{
    \begin{tabular}{ccccccccc}
        \toprule
         \textbf{PEFT Method} & \textbf{\# Params (\%)} & \textbf{ST/MT} & \textbf{ARC-e} & \textbf{ARC-c} & \textbf{BoolQ} & \textbf{OBQA} & \textbf{PIQA} & \textbf{AVG.} \\ 
        \midrule
        \multirow{2}{*}{LoRA}  & 3.2\% & ST & 71.9  & 43.2  & 62.1  & 71.4  & 80.9  & 65.9  \\ 
        ~ & 3.2\% & \textbf{MT} & 64.9  & 50.2  & 66.4  & 64.8  & 75.7  & 64.4  \\ 
        \rowcolor{Gray} 
        ~ & ~ & ~ & \textcolor{myred}{-7.0} & \textcolor{mygreen}{7.0} & \textcolor{mygreen}{4.3} & \textcolor{myred}{-6.6} & \textcolor{myred}{-5.2} & \textcolor{myred}{-1.5} \\
        \midrule
        \multirow{2}{*}{DoRA} & 3.2\% & ST & 71.5  & 46.2  & 62.2  & 70.4  & 81.6  & 66.4  \\ 
        ~ & 3.2\% & \textbf{MT} & 63.7  & 50.8  & 61.6  & 61.0  & 81.1  & 63.6  \\ 
        \rowcolor{Gray} 
         ~ & ~ & ~ &  \textcolor{myred}{-7.8} & \textcolor{mygreen}{4.6} & \textcolor{myred}{-0.6} & \textcolor{myred}{-9.4} & \textcolor{myred}{-0.5} & \textcolor{myred}{-2.8} \\ 
        \midrule
        \multirow{2}{*}{\textbf{MixLoRA}} & 4.3\% & ST & 76.3  & 47.4  & 65.8  & 75.8  & 81.1  & 69.3  \\ 
        ~ & 4.3\% & \textbf{MT} & 70.3  & 55.5  & 66.6  & 70.0  & 78.7  & 68.2  \\
        \rowcolor{Gray} 
        ~ & ~ & ~ &   \textcolor{myred}{-6.0} & \textcolor{mygreen}{8.1} & \textcolor{mygreen}{0.8} & \textcolor{myred}{-5.8} & \textcolor{myred}{-2.4} & \textcolor{myred}{-1.1} \\ 
        \midrule
        \multirow{2}{*}{\textbf{MixDoRA}} & 4.3\% & ST & 77.0  & 54.3  & 67.2  & 75.4 & 81.8  & 71.1  \\ 
        ~ & 4.3\% & \textbf{MT} & 71.1  & 56.3  & 65.9  & 70.6  & 79.1  & 68.6 \\ 
        \rowcolor{Gray} 
        ~ & ~ & ~ &  \textcolor{myred}{-5.9} & \textcolor{mygreen}{2.0} & \textcolor{myred}{-1.3} & \textcolor{myred}{-4.8} & \textcolor{myred}{-2.7} & \textcolor{myred}{-2.5} \\ 
        \bottomrule
    \end{tabular}}
    \vspace{2mm}
\label{table:multi_task_results_gemma}
\end{table}


\subsection{Experimental Results of Performance Metrics. } \label{appd:efficiency}

\begin{table}[h]
\renewcommand\arraystretch{1.25}
\centering
\caption{Experimental results of LLaMA-2 7B for performance metrics. The latency shown in the table represents the token computation latency, and the memory indicates the peak GPU memory collected by the profiler. To accurately measure the performance of LoRA and DoRA during inference, we conducted the experiments with weights unmerged. \dagnote{} represents methods with \model{} optimization.}
\vspace{2mm}
\scalebox{0.85}{
\begin{tabular}{cccccccccccc}
 \toprule
 \multirow{3}*{\textbf{PEFT Method}}
 & \multicolumn{6}{c}{\textit{Training}} 
 & \multicolumn{4}{c}{\textit{Inference}} \\
 \cline{2-11}
 & \multicolumn{2}{c}{\textbf{Forward}}
 & \multicolumn{2}{c}{\textbf{Backward}}
 & \multicolumn{2}{c}{\textbf{Memory}} 
 & \multicolumn{2}{c}{\textbf{Forward}}
 & \multicolumn{2}{c}{\textbf{Memory}} \\
 \cline{2-11}
 & $\mu s$ & \% & $\mu s$ & \% & GB & \% & $\mu s$ & \% & GB & \% \\
 \hline
 LoRA
 & 245.3 & 100.0\% & 552.3  & 100.0\% & 15.2 & 100.0\% & 241.4 & 100.0\% & 13.7 & 100.0\% \\
 \rowcolor{Gray} 
 DoRA
 & 659.4 & 268.8\% & 1193.8 & 216.1\% & 15.6 & 102.4\% & 645.3 & 267.3\% & 13.7 & 100.0\% \\
 \hline
 MixLoRA
 & 535.2 & 218.2\% & 1187.5 & 215.0\% & 15.1 & 99.5\% & 532.8 & 220.7\% & 13.7 & 100.0\% \\
 \rowcolor{Gray} 
 \dagnote{}MixLoRA
 & 462.5 & 188.5\% & 1097.6 & 198.7\% & 15.1 & 99.5\% & 442.2 & 183.2\% & 13.7 & 100.0\% \\
 \hline
 MixLoRA $\times 2$           
 & 533.9 & 217.7\% & 1185.5 & 214.6\% & 8.8 & 57.7\% & 523.8 & 217.0\% & 7.2 & 52.5\% \\
 \rowcolor{Gray} 
 \dagnote{}MixLoRA $\times 2$
 & 441.0 & 179.8\% & 1072.3 & 194.1\% & 8.8 & 57.7\% & 441.4 & 182.8\% & 7.2 & 52.5\% \\
 \bottomrule
\end{tabular}}
\label{table:performance_results_llama}
\end{table}

\begin{table}[h]
\renewcommand\arraystretch{1.25}
\centering
\caption{Experimental results of Gemma 2B for performance metrics. The latency shown in the table represents the token computation latency, and the memory indicates the peak GPU memory collected by the profiler. To accurately measure the performance of LoRA and DoRA during inference, we conducted the experiments with weights unmerged. \dagnote{} represents methods with \model{} optimization.}
\vspace{2mm}
\scalebox{0.85}{
\begin{tabular}{cccccccccccc}
 \toprule
 \multirow{3}*{\textbf{PEFT Method}}
 & \multicolumn{6}{c}{\textit{Training}} 
 & \multicolumn{4}{c}{\textit{Inference}} \\
 \cline{2-11}
 & \multicolumn{2}{c}{\textbf{Forward}}
 & \multicolumn{2}{c}{\textbf{Backward}}
 & \multicolumn{2}{c}{\textbf{Memory}} 
 & \multicolumn{2}{c}{\textbf{Forward}}
 & \multicolumn{2}{c}{\textbf{Memory}} \\
 \cline{2-11}
 & $\mu s$ & \% & $\mu s$ & \% & GB & \% & $\mu s$ & \% & GB & \% \\
 \hline
 LoRA
 & 151.1 & 100.0\% & 308.2  & 100.0\% & 11.4 & 100.0\% & 152.0 & 100.0\% & 10.6 & 100.0\% \\
 \rowcolor{Gray} 
 DoRA
 & 539.4 & 356.9\% & 919.9 & 298.5\% & 11.4 & 100.0\% & 533.4 & 350.9\% & 10.6 & 100.0\% \\
 \hline
 MixLoRA
 & 250.6 & 165.8\% & 527.2 & 171.1\% & 11.2 & 97.7\% & 245.1 & 161.2\% & 10.5 & 99.8\% \\
 \rowcolor{Gray} 
 \dagnote{}MixLoRA
 & 226.5 & 149.9\% & 525.2 & 170.4\% & 11.2 & 97.7\% & 224.0 & 147.4\% & 10.5 & 99.8\% \\
 \hline
 MixLoRA $\times 2$           
 & 249.6 & 165.1\% & 524.0 & 170.1\% & 7.6 & 66.9\% & 243.1 & 160.6\% & 6.5 & 61.5\% \\
 \rowcolor{Gray} 
 \dagnote{}MixLoRA $\times 2$
 & 223.8 & 148.1\% & 523.7 & 169.9\% & 7.6 & 66.9\% & 221.2 & 145.6\% & 6.5 & 61.5\% \\
 \bottomrule
\end{tabular}}
\label{table:performance_results_gemma}
\end{table}

Table~\ref{table:performance_results_llama} shows the results on LLaMA2 7B, demonstrating that \model{} exhibits lower token computation latency (DoRA requires 659.4$\mu$s for forward propagation, while \model{} only needs 535.2$\mu$s) and comparable peak GPU memory usage to DoRA (approximately 15GB). However, \model{} shows nearly twice the token computation latency of LoRA (245.3$\mu$s). This increased latency is due to \model{} sending each token to two experts for computation (when $K = 2$). Nonetheless, with our optimized algorithm, we reduced the token computation latency by nearly 30\% for a single model (from 535.2$\mu$s to 462.5$\mu$s) and decreased the peak GPU memory per model by almost 45\% when training or inferring with two models simultaneously (from 15.1GB to 8.8GB during training, and from 13.7GB to 7.2GB during inference). Appendix~\ref{table:performance_results_gemma} shows the results on Gemma 2B, corroborating these findings and proving that our algorithm maintains robustness across different model sizes. In conclusion, experiments show that \model{} offers a more balanced trade-off, providing higher performance with reduced latency compared to the current state-of-the-art method, DoRA.

\newpage
\subsection{Robustness of \model{} Towards Different Rank}  \label{appd:Rank}
\begin{table}[h]
\caption{Accuracy comparison of \model{} and \doramodel{} with varying ranks for LLaMA2-7B on the commonsense reasoning tasks.}
\label{table:Robustness_Rank}
\centering
\begin{tabular}{cccccccc}
\toprule
\textbf{PEFT Method}      & \textbf{Rank $r$} & \textbf{\# Params (\%)}  & \textbf{ARC-e} & \textbf{ARC-c} & \textbf{BoolQ} & \textbf{OBQA} & \textbf{Avg.} \\
\midrule
\multirow{5}{*}{\textbf{MixLoRA}}
      & 2 &{0.38\%} &76.1 &56.4 &73.3 &79.2 &71.3 \\ 
    ~ &4 &{0.74\%} &76.2 &56.5 &73.8 &80.8 &71.8\\
    ~ &8 &{1.46\%} &76.9 &56.8 &74.2 &81.2 &72.3\\
    ~ &16&{2.91\%} &77.7 &58.1 &72.7 &84.4 &\textbf{73.2}\\
    ~ &32&{5.80\%} &79.1 &54.1 &70.0 &76.4 &69.9 \\
\midrule
\multirow{5}{*}{\textbf{MixDoRA}}
      &2 &{0.38\%} &75.3 &52.7 &73.3 &80.4 &70.4 \\ 
    ~ &4 &{0.74\%} &76.2 &55.0 &73.2 &80.4 &71.2\\
    ~ &8 &{1.46\%} &76.7 &55.5 &73.5 &78.6 &71.1\\
    ~ &16&{2.91\%} &77.5 &58.2 &72.6 &80.9 &\textbf{72.3}\\
    ~ &32&{5.80\%} &75.5 &53.6 &72.0 &77.6 &69.7 \\
\bottomrule
\end{tabular}
\end{table}

\subsection{Robustness of \model{} Towards Different Router Loss Coefficient} \label{appd:Loss}
\begin{table}[h]
\caption{Accuracy comparison of \model{} and \doramodel{} with different Router Loss for LLaMA2-7B on the commonsense reasoning tasks.}
\label{table:Robustness_Router_Loss_Coef}
\centering
\begin{tabular}{ccccccc}
\toprule
\textbf{PEFT Method} & \textbf{Router Loss Coef.} & \textbf{ARC-e} & \textbf{ARC-c} & \textbf{BoolQ} & \textbf{OBQA} & \textbf{Avg.} \\
\midrule
\multirow{4}{*}{\textbf{MixLoRA}}
      & -    & 75.5 & 55.5 & 72.8 & 78.8 & 70.5 \\ 
    ~ & 1e-3 & 77.7 & 58.1 & 72.7 & 84.4 & \textbf{73.2} \\
    ~ & 1e-2 & 77.0 & 56.4 & 73.1 & 80.6 & 71.8 \\
    ~ & 1e-1 & 76.6 & 55.7 & 72.7 & 80.8 & 71.5 \\
\midrule
\multirow{4}{*}{\textbf{MixDoRA}} 
      & -    &77.7 &56.9 &72.8 &79.2 & 71.5 \\ 
    ~ & 1e-3 &77.5 &58.2 &72.6 &80.9 & \textbf{72.3} \\
    ~ & 1e-2 &77.6 &56.2 &73.0 &80.6 & 71.9 \\
    ~ & 1e-1 &77.3 &54.6 &72.0 &79.8 & 70.9 \\
\bottomrule
\end{tabular}
\end{table}

\clearpage

\subsection{Optimization Algorithm} \label{appd:optimization}

\begin{wrapfigure}{r}{0.6\textwidth}
 \vspace{-2mm}
  \begin{minipage}{0.6\textwidth}
  \vspace{-6mm}
\begin{algorithm}[H]
\caption{Optimal Forward Propagation of \model{}}
\small
\centering  
\begin{algorithmic}[1]
\REQUIRE{multi-task token sequence $\mathbf{T}^{l-1}$: \textcolor{shapecolor}{$(\mathtt{M}, \mathtt{B}, \mathtt{N}, \mathtt{D})$}}
\ENSURE{multi-task token sequence $\mathbf{T}^{l}$: \textcolor{shapecolor}{$(\mathtt{M}, \mathtt{B}, \mathtt{N}, \mathtt{D})$}}
\STATE \textcolor{gray}{\text{/* Allocate multi-task sequences to various \model{}s */}}
\FOR{$t$ in \{multi-task sequences $\mathbf{T}^{l-1}$\}}
    \STATE $\mathbf{T}_{t}^{l-1}$: \textcolor{shapecolor}{$(\mathtt{B}, \mathtt{N}, \mathtt{D})$} $\leftarrow$ $\mathbf{T}^{l-1}[t, :, :, :]$
    \STATE $\mathbf{r}_{t}$: \textcolor{shapecolor}{$(\mathtt{B} \times \mathtt{N}, \mathtt{K})$} $\leftarrow$ $\mathbf{Linear}_{t}(\mathbf{T}_{t}^{l-1})$
    \STATE $\mathbf{r}'_{t}$: \textcolor{shapecolor}{$(\mathtt{B} \times \mathtt{N}, \mathtt{K})$} $\leftarrow$ $\mathbf{Norm}(\mathbf{Top2}(\mathbf{Softmax}(\mathbf{r}_{t})))$
    \STATE \textcolor{gray}{\text{/* Reduction of duplicative calculations */}}
    \STATE $\bar{\mathbf{h}}^{W_1}_{t}$: \textcolor{shapecolor}{$(\mathtt{B}, \mathtt{N}, \mathtt{D'})$} $\leftarrow$ $\mathbf{Linear}^{W_1}_{l}(\mathbf{T}_{t}^{l-1})$
    \STATE $\bar{\mathbf{h}}^{W_3}_{t}$: \textcolor{shapecolor}{$(\mathtt{B}, \mathtt{N}, \mathtt{D'})$} $\leftarrow$ $\mathbf{Linear}^{W_3}_{l}(\mathbf{T}_{t}^{l-1})$
    \STATE \textcolor{gray}{\text{/* Calculate the outputs of LoRA experts */}}
    \FOR{$k$ in \{LoRA experts $\mathbf{E}^{l}_{t}$\}}
        \STATE $\hat{\mathbf{h}}^{W_1}_{t}$: \textcolor{shapecolor}{$(\mathtt{B}, \mathtt{N}, \mathtt{D'})$} $\leftarrow$ $\bar{\mathbf{h}}^{W_1}_{t} + \mathbf{LoRA}^{W_1}_{k}(\mathbf{T}_{t}^{l-1})$
        \STATE $\hat{\mathbf{h}}^{W_3}_{t}$: \textcolor{shapecolor}{$(\mathtt{B}, \mathtt{N}, \mathtt{D'})$} $\leftarrow$ $\bar{\mathbf{h}}^{W_3}_{t} + \mathbf{LoRA}^{W_3}_{k}(\mathbf{T}_{t}^{l-1})$
        \STATE $\hat{\mathbf{h}}_{t}$: \textcolor{shapecolor}{$(\mathtt{B}, \mathtt{N}, \mathtt{D'})$} $\leftarrow$ $\mathbf{SiLU}(\hat{\mathbf{h}}^{W_1}_{t}) \bigodot \hat{\mathbf{h}}^{W_3}_{t}$
        \STATE $\mathbf{h}_{t}$: \textcolor{shapecolor}{$(\mathtt{B}, \mathtt{N}, \mathtt{D})$} $\leftarrow$ $\mathbf{Linear}^{W_2}_{l}(\hat{\mathbf{h}}_{t}) + \mathbf{LoRA}^{W_2}_{k}(\hat{\mathbf{h}}_{t})$
        \STATE $\mathbf{T}^{l}_{t}$: \textcolor{shapecolor}{$(\mathtt{B}, \mathtt{N}, \mathtt{D})$} $\leftarrow$ $\mathbf{T}^{l}_{t} + \mathbf{h}_{t} \bigodot \mathbf{r}'_{t}[:, k]$
    \ENDFOR
    \STATE $\mathbf{T}^{l}$: \textcolor{shapecolor}{$(\mathtt{t+1}, \mathtt{B}, \mathtt{N}, \mathtt{D})$} $\leftarrow$ $\mathbf{concat}(\mathbf{T}^{l}, \mathbf{T}^{l}_{t}.\mathbf{unsqueeze}(0))$
\ENDFOR
Return: $\mathbf{T}^{l}$ 
\label{alg:mixlora}
\end{algorithmic}
\end{algorithm}
  \end{minipage}
\end{wrapfigure}

By combining two optimization strategies mentioned in Section~\ref{sec:optimization}, reducing computational complexity~(I) and multi-model high-throughout training~(II), we propose a forward propagation algorithm described in Algorithm 1. Specifically, the multi-task input sequences $\mathbf{T}^{l-1}$ include various token sequences from $M$ tasks, where each token sequence is sequentially allocated to different \model{} modules for processing (line 1). Given that the pretrained dense model weights remain frozen, it becomes feasible to maintain two or more \model{} models that share the same pretrained dense model weights. This approach reduces GPU memory cost and improves training efficiency by allowing multiple \model{} modules to be trained on a single GPU and reducing kernel launch time.
Next, we linearly project the token sequences of task $t$ to the logits $\mathbf{r}_{t}$ (line 4) and compute the normalized logits $\mathbf{r}'_{t}$ of activated experts by employing Softmax and Top-2 functions (line 5). We observe that the shared FFN sublayer repeatedly participates in the computation in multiple LoRA experts, which can be avoided with the linear layer $W_1$ and the linear layer $W_3$ of the FFN. Therefore, the projected token sequences $\bar{\mathbf{h}}^{W_1}_{t}$ and $\bar{\mathbf{h}}^{W_3}_{t}$ are saved in advance before computing the outputs of each expert (lines 7 and 8). Finally, we compute the product of the $k$-th LoRA metrics $LoRA_{k}$ plus the shared FFN weights as the weights of the $k$-th expert and weight the outputs of all activated experts with the logits $\mathbf{r}'_{t}$ generated by the router to get the output token sequence of the $t$-th task (line 15).

\end{document}